\documentclass[journal]{IEEEtran}
\usepackage[pdftex]{graphicx}
\usepackage{amsmath}
\usepackage{algorithmic}
\usepackage{color,soul}
\usepackage{array}
\usepackage{url}
\usepackage{svg}
\usepackage{graphicx}
\usepackage{lscape}
\usepackage[font=large]{subfig}
\usepackage{blindtext}
\usepackage{amssymb}
\usepackage{scrextend}
\usepackage{tikz}
\usetikzlibrary{shadows.blur}
\usepackage[edges]{forest}
\usepackage{hyperref}
\usepackage{booktabs}
\usepackage{multirow}
\usepackage{footnote}
\usepackage{mathtools}
\usepackage{chemformula}
\usepackage[usestackEOL]{stackengine}
\usepackage{footnote}
\makesavenoteenv{tabular}
\makesavenoteenv{table}
\usepackage[flushleft]{threeparttable}
\usepackage{adjustbox}
\usepackage{enumitem}
\usepackage{flushend}

\begin{document}

\title{Deep Learning for Road Traffic Forecasting:\\Does it Make a Difference?}

\author{Eric L. Manibardo,
        Ibai La\~na,
        and Javier Del Ser,~\IEEEmembership{Senior Member,~IEEE}
        \thanks{Eric L. Manibardo, Ibai La\~na and Javier Del Ser are with TECNALIA, Basque Research and Technology Alliance (BRTA), 48160 Derio, Bizkaia, Spain.
        Javier Del Ser is also with the University of the Basque Country (UPV/EHU) 48013 Bilbao, Bizkaia, Spain}
        \thanks{e-mail: [eric.lopez, ibai.lana, javier.delser]@tecnalia.com}
        }


%
\maketitle

\begin{abstract}
Deep Learning methods have been proven to be flexible to model complex phenomena. This has also been the case of Intelligent Transportation Systems (ITS), in which several areas such as vehicular perception and traffic analysis have widely embraced Deep Learning as a core modeling technology. Particularly in short-term traffic forecasting, the capability of Deep Learning to deliver good results has generated a prevalent inertia towards using Deep Learning models, without examining in depth their benefits and downsides. This paper focuses on critically analyzing the state of the art in what refers to the use of Deep Learning for this particular ITS research area. To this end, we elaborate on the findings distilled from a review of publications from recent years, based on two taxonomic criteria. A posterior critical analysis is held to formulate questions and trigger a necessary debate about the issues of Deep Learning for traffic forecasting. The study is completed with a benchmark of diverse short-term traffic forecasting methods over traffic datasets of different nature, aimed to cover a wide spectrum of possible scenarios. Our experimentation reveals that Deep Learning could not be the best modeling technique for every case, which unveils some caveats unconsidered to date that should be addressed by the community in prospective studies. These insights reveal new challenges and research opportunities in road traffic forecasting, which are enumerated and discussed thoroughly, with the intention of inspiring and guiding future research efforts in this field.
\end{abstract}


\section{Introduction} \label{sec:intro}

It is undeniable that the boom of the Big Data era has revolutionized most research fields \cite{suchanek2013knowledge}. The reason for this advent is that much more data are collected from a variety of sources, which must be processed and converted into various forms of knowledge for different stakeholders. Intelligent Transportation Systems (ITS), which aim to improve efficiency and security of transportation networks, embody one of the domains that has largely taken advantage of the availability of data generated by different processes and agents that interact with transportation. Some examples of ITS applications and use cases that benefit from data availability are railway passenger train delay prediction \cite{yaghini2013railway}, airport gate assignment problem \cite{xu2001airport}, adaptive control of traffic signaling in urban areas \cite{mannion2016experimental} and improvements of autonomous driving \cite{muhammad2020deep}, to mention a few.

Within the diversity of ITS sub-domains, this work is focused on traffic state forecasting. An accurate traffic state prediction, based on measurements of different nature (e.g. average speed, occupancy, travel time, etc.), can be used to enhance traffic management and implement operational measures to relieve or prevent traffic congestion and its consequent implications \cite{litman2009transportation,levy2010evaluation}. Motivated by this problem, a plethora of short-term traffic forecasting works is published every year, as can be seen in recent surveys on this topic \cite{vlahogianni2004short, vlahogianni2014short, lana2018road}.

Although there are plenty of data-driven methods that can deliver a short-term traffic prediction model, Deep Learning methods have monopolized the majority of publications of this type in recent years, becoming the reference for the community when facing new forecasting problems \cite{zhao2017lstm,polson2017deep}. This apogee of the application of Deep Learning methods to ITS problems is commonly justified by its theoretical capability to approximate any non-linear function \cite{saxe2013exact}, which is often the case of patterns underneath traffic time series \cite{nair2001non}. In general, short-term prediction models estimate future time series values based on recent measurements, whereas long-term traffic forecasting solutions rather focuses on finding typical traffic profiles. However, Deep Learning models have their own drawbacks in the form of an inability to understand their behavior \cite{gunning2017explainable, arrieta2020explainable}, and the need for large quantities of data and specialized hardware resources.

Under this premise, this work elaborates on Deep Learning for short-term traffic forecasting in order to ascertain the areas in which its implementation brings the best outcomes, as well as other scenarios where less computational expensive data-driven methods provide similar or superior performance. To shed light on this matter, we first analyze thoroughly recent literature on traffic forecasting, specifically those works that propose Deep Learning based solutions. Parting from this prior analysis of the state of the art, we enumerate and discuss a series of insights, good and poor practices followed by the community to date. Our critical analysis is supported by the results yielded by an experimental study comprising several shallow and Deep Learning models and traffic forecasting datasets. Finally, we outline several research niches, open challenges, and valuable research directions for the community, in close connection to the overall conclusions drawn from our study on the current status of the field. In summary, the main contributions of this paper can be described as follows:
\begin{itemize}[leftmargin=*]
  \item We categorize recent Deep Learning based short-term traffic forecasting publications under a taxonomy that considers two criteria: 1) the specific forecasting problem; and 2) the selected techniques and methods to model the actual phenomena.
  \item We critically examine the state of the art by following the above criteria, which allow us to detect research trends and to extract insights about overlooked issues and pitfalls.
  \item We design an extensive experimentation comprising traffic data of different characteristics (e.g. highways and urban arterials) captured from several locations, covering the most common scopes of traffic forecasting, intending to show good practices when evaluating the performance of novel traffic forecasting techniques. 
  \item We enumerate a series of learned lessons drawn from the experimental setup, underscoring poor research habits that should be avoided for the sake of valuable advances in the field.
  \item Finally, we discuss challenges and research opportunities of the field, all directed towards achieving actionable and trustworthy short-term traffic forecasting models.
\end{itemize}

The rest of the paper is organized as follows: Section \ref{sec:concepts} provides an introduction to the evolution of short-term traffic forecasting in recent years, Deep Learning concepts, and how this technology has become the spearhead of the traffic prediction field. Section \ref{sec:review+taxonomy} defines the proposed taxonomic criteria, classifies and reviews the recent state of the art. A discussion on the findings and conclusions drawn by the previous review is held in Section \ref{sec:criticalanalysis}. Next, a case study is conducted in Section \ref{sec:casestudy}, and lessons learned therefrom are covered in Section \ref{sec:learnedlessons}. Challenges and research opportunities related to short-term traffic forecasting are discussed in Section \ref{sec:challenges}. Finally, Section \ref{sec:conclusions} ends this survey with a summary of final thoughts and an outlook.

\section{Concepts and Preliminaries}\label{sec:concepts}

Short-term traffic forecasting has been one of the cornerstones for traffic management, as it is a reliable tool to manage and maintain traffic networks. In turn, Deep Learning comprehends a mixture of data-driven models which excellent results in many applications have stimulated their widespread adoption for short-term traffic forecasting. With that in mind, the trajectory of both research fields and their relationships are reviewed in this section, in order to provide a better understanding of how Deep Learning techniques have become dominant in the short-term traffic forecasting realm.

\subsection{Deep Learning}\label{subsec:deep}

Machine Learning techniques provide a compendium of tools to develop data-based mathematical representations of real-world processes. These representations allow automatizing certain tasks or even predicting future states of the processes being modeled. As a subset of Machine Learning, Deep Learning is inspired by the structure of human brains. The hierarchical composition of neural units, which are the fundamental building block of Deep Learning architectures, allows theoretically approximating any kind of non-linear function \cite{lecun2015deep}. Since in nature there is an abundance of processes that can be modeled as non-linear functions, Deep Learning has quickly become the dominant approach in many applications. The capabilities of Deep Learning have been particularly relevant in natural language processing \cite{otter2020survey} and computer vision \cite{liu2020deep}, among others, revolutionizing those fields. As a consequence, scholars are constantly applying these techniques to other areas of knowledge, seeking to extrapolate the benefits observed for these applications to other domains.

Deep Learning models, like others belonging to different subsets of Machine Learning, can perform many tasks such as unsupervised learning, classification, or regression. But what makes them particularly relevant is their unique capabilities to automatically learn hierarchical features from data that are useful for the task under consideration. Classical Machine Learning methods are also called \emph{flat} or \emph{shallow learning} methods because they cannot learn data representations directly from unprocessed data. Feature extraction needs to be applied beforehand, often assisted by expert knowledge about the domain of the problem. Deep Learning methods, however, can learn an implicit representation of raw data for a better understanding of the process to be modeled. This capability has been proven to go beyond human reasoning limits. As a result, for many fields dealing with complex, highly dimensional data, features discovered by Deep Learning methods lead to unprecedented performance with respect to the state of the art.

The other main capability of Deep Learning methods stems from their architectural flexibility: data fusion. Deep Learning flexible architectures allow for the different format data types to be merged, combining the information of multiple sources and extracting more knowledge about the process to model. Therefore, Deep Learning allows researchers to resolve complex learning problems, specially when dealing with highly-dimensional data.  
\begin{figure*}[t!]
    \centering
    \includegraphics[width=\linewidth]{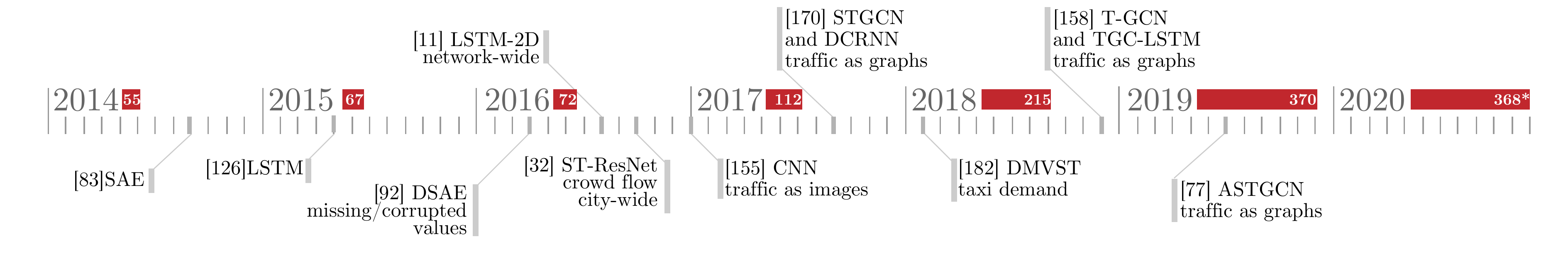}
    \caption{Milestones in the history of Deep Learning based short-term traffic forecasting. Publications are ordered according to their publication date. Horizontal red bars denote the number of works that were published every year concerning this topic (retrieved from Scopus in November 2020 with query terms: [\texttt{neural network} OR  \texttt{deep learning} OR \texttt{deep neural network} OR \texttt{LSTM network} OR \texttt{deep spatio-temporal}] AND [\texttt{traffic prediction} OR \texttt{traffic forecasting} OR \texttt{traffic congestion} OR \texttt{traffic state prediction}], in title, abstract or keywords).}
    \label{fig:cronologia}
\end{figure*}

\subsection{Short-term traffic forecasting}

The development of the short-term traffic forecasting field began when researchers started to apply time series forecasting methods to characterize traffic congestion measurements \cite{ahmed1979analysis}. Back then, one popular approach relied on the assumption that the process that generated the traffic time series could be approximated using statistical methods like auto-regressive integrated moving average (ARIMA) \cite{levin1980forecasting,kawashima1987long}. These predictive models were only capable of predicting a single target point of a road map.

With the beginning of the new millennium, the complexity of modeling techniques started to increase sharply, unleashing new research opportunities for the traffic forecasting arena. Vlahogianni et al. \cite{vlahogianni2014short}, who analyzed short-term forecasting literature from 2004 to 2012, brought up that researchers are distancing themselves from what are considered classical statistical methods (i.e. auto-regressive models), drifting towards data-driven approaches \cite{karlaftis2011statistical}. The primary motivation for this shift remains on the ineffectiveness of classical methods to forecast while facing unstable conditions. The nature of the traffic is not stationary or linear, as a manifold of studies have hitherto shown \cite{vlahogianni2006statistical, vlahogianni2011temporal, kamarianakis2010characterizing,tang2014dynamic}. Unfortunately, auto-regressive models tend to focus on the average behavior, so peaks and rapid fluctuations are generally missed \cite{vlahogianni2004short}. Further into the review in \cite{vlahogianni2014short}, the literature analyzed therein inspected the scope of application, input and output data type, prediction horizon, and proposed technique of publications. Finally, challenges identified in this seminal review stressed out the overabundance of studies focused on freeways and interurban road traffic. Models for urban road traffic data were revealed to be less frequently studied. Furthermore, only a few solutions capable of predicting traffic simultaneously at different locations of the road network were known at the time \cite{cheng2012spatio, kamarianakis2012real, sun2012network}, due to the scarcity of open-access traffic data for numerous points in a network, together with the high complexity of solving the interactions between the studied roads of the area.

After assimilating the criticism and challenges established in \cite{vlahogianni2014short}, another survey \cite{lana2018road}, published years thereafter, proposed new insights unattended until then. The newer literature review over the 2014-2016 period showed an increase in the number of publications focused on prediction at urban roads, which evinced that the research field covers nowadays most of possible geographic contexts of traffic prediction. Also in connection with the prospects in \cite{lana2018road}, there is also an increasing interest within the community in obtaining network-wide predictions, possibly promoted by the improvement in spatial data coverage and computing capacity achieved over the years \cite{ma2015large,zhang2016dnn}. 

Among other points, \cite{lana2018road} also underscored the need for establishing a unified set of metrics that permit to fairly compare performance between different models. Absolute error metrics provide interpretable values when comparing models for a same dataset, enabling a qualitative analysis of the error, as these express the error into traffic units (for instance, vehicles per hour). However, if the benchmark comprises several traffic datasets, relative error metrics should be considered for proper model comparison. This way the magnitude of the traffic unit does not affect the comparison study. Lastly, this survey highlighted an intrinsic problem of data-driven models: concept drift \cite{gama2014survey}. Since data-driven models acquire information from large data collections in order to extract traffic patterns and provide accurate predictions, performance is affected by exogenous non-planned events such as accidents, roads works or other circumstantial changes.

That same year, Ermagun et al. \cite{ermagun2018spatiotemporal} analyzed the methodology and proposed methods for capturing spatial information over road networks. Their assumption is that present information of spatial relationships between road nodes should improve short-term predictive model performance. The study, which spans the period 1984-2016, offers an overview of the concerns of researchers in the field: 65.3\% of revised works are concentrated on traffic flow, 19.2\% speed, and the remaining travel time. Likewise, only 26.5\% chose urban zones as the implementation area, whereas the remainder are concentrated at freeways, confirming the postulated trend of Vlahogianni et al. in \cite{vlahogianni2014short}. Finally, the survey concludes by encouraging the community to portray road networks as graphs \cite{rodrigue2016geography}, since they ease the representation of inter-nodal relationships and their subsequent use in modeling.

To round up this tour on the recent history of the field, in 2019 Angarita et al. \cite{angarita2019taxonomy} propose a general taxonomy for traffic forecasting data-driven models. The motivation of their work is not only to classify and revise learning models used to date, but also to categorize the approached traffic forecasting problems. In terms of data source type, data granularity, input and output nature, and overall scope. On the other hand, the reviewed models are sorted by pre-processing technique, type of in/out data, and step-ahead prediction. After analyzing the state of the art, they find no data-driven approach that suits all forecasting situations.

All the above surveys offer insights into the goals pursued by the field, as well as an outline of the opportunities and challenges that should be addressed in prospective studies. Vlahogianni et al. advocate for data-driven approaches, which were already gaining impulse at the time \cite{vlahogianni2014short}. Posterior surveys confirmed this trend, and data-driven models prevail nowadays as the preferred option for short-term traffic modeling. The work of La\~{n}a et al. concludes that most possible geographic scopes are covered in the state of the art since, in the origins of the short-term traffic forecasting field, there was a shortage of publications based on urban traffic data \cite{lana2018road}. In turn, Ermagun et al. grant importance to spatio-temporal relationships between nodes of traffic networks, which is one of the most exploited relationships to extract knowledge in the actual literature \cite{ermagun2018spatiotemporal}. On a closing note, the taxonomy of Angarita et al. in \cite{angarita2019taxonomy} classifies traffic forecasting publications from a supervised learning perspective, which inspires in part the criteria later adopted in this work.

\subsection{When Deep Learning meets traffic forecasting}

Table \ref{tab: surveys} summarizes the criteria under consideration for each survey that has been published so far on Deep Learning models for short-term traffic forecasting. 
\begin{table*}[ht]
\centering
    \begin{adjustbox}{max width=\textwidth}
    \begin{threeparttable}
        \caption{Published surveys that address short-term traffic forecasting based on Deep Learning methods. Column headers denotes the citation reference of each publication. Rows correspond to different characteristics and content included in each survey. }
        \label{tab: surveys}
        \begin{tabular}{lccccccccc} 
         \toprule
         \multirow{1}{*}{\textbf{Survey}}  &
         
         \multicolumn{1}{c}{\cite{do2019survey}} &
         \multicolumn{1}{c}{\cite{nguyen2018deep}}  &
         \multicolumn{1}{c}{\cite{wang2019enhancing}} &
         \multicolumn{1}{c}{\cite{wang2020deep}}   &
         \multicolumn{1}{c}{\cite{yin2020comprehensive}} &
         \multicolumn{1}{c}{\cite{tedjopurnomo2020survey}}   &
         \multicolumn{1}{c}{\cite{gobezie2020machine}} &
         \multicolumn{1}{c}{\cite{lee2020short}} &
         \multicolumn{1}{c}{\textbf{Ours}}  \\
         
         \cmidrule(lr){1-10}
         
         \textbf{Period}                 & 1994 - 2018 & 1997 - 2017   & 1999 - 2018 & 2012 - 2018 & 2015 - 2020& 2014-2019  & 2014 - 2019 & 2014 - 2020 & 2015 - 2020\\
         \textbf{\# of reviewed works}   & $\sim70$ & $\sim15$  & $\sim35$ & $\sim80$ &$\sim65$ & $\sim40$  & $\sim10$  & $\sim100$ & $\sim150$\\
         \textbf{Measurement}            &  F, S, O & F, D     & F, S, TT & - &  F, S, D, O, TT & F, S   & F & F, S, D, C, A & F, S, D, O, TT \\
         \textbf{Context}                & U, H  & -    & - & - & - & - & - & - & U, H\\
         \textbf{Sensing technique}      & - & -    & -   & -  & - &- & Yes  & - & Yes\\
         \textbf{Temporal resolution}    & Yes & -    & -   & - & - & Yes  &- & - & Yes\\
         \textbf{Dependencies}           & ST, T & -       & - & ST &- & ST, T & - & ST, T & ST, T\\
         \textbf{Image representation}   & - & - & - & - & - & - &-& - & Yes\\
         \textbf{Graph representation}   & - & - & - & Yes & Yes & -  & - & Yes & Yes\\
         \textbf{Coverage}               & - & -    & -   & - & - & - & - & - & Yes\\
         \textbf{\# of steps ahead}      & Yes & -    & -   & - & - &- & - & - & Yes\\
         \textbf{Model type}             & Yes  & Yes  & Yes   & Yes & Yes &Yes & Yes & Yes & -\\
         \textbf{Empirical study}        & -  & -  & -   & - & Yes & -  & - & - & Yes \\
         
         \bottomrule
         
        \end{tabular}
        \begin{tablenotes}
          \item Note: The row "\# of reviewed works" only takes into account publications related to short-term traffic forecasting based on Deep Learning methods. Any other unrelated reference has been filtered out and not accounted for in the reported quantities. F: Flow; S: Speed; D: Demand; O: Occupancy; TT: Travel Time; A: Accidents U: Urban; H: Highways/Freeways; ST: Spatio-temporal; T: Temporal.
        \end{tablenotes}
  \end{threeparttable}
  \end{adjustbox}
\end{table*}

As it can be concluded from the most recent surveys on short-term traffic forecasting, Deep Learning models have been applied in this research area mostly since the last decade. Figure \ref{fig:cronologia} depicts a timeline with important milestones and achievements in short-term traffic forecasting approached via Deep Learning models. Among them, recent surveys that address short-term traffic forecasting in conjunction with Deep Learning methods are analyzed in this section, in order to highlight the need for the synthesis and investigation presented in this work.

Starting with \cite{do2019survey}, this work focuses on different Deep Learning architectures applied for short-term traffic forecasting and explains their components and operation. A categorization of the reviewed models is presented, providing an overview of new modeling proposals. The second and third surveys \cite{nguyen2018deep,wang2019enhancing} analyze several Deep Learning methods for different transportation topics, including traffic signal control, autonomous driving and traffic state prediction. Therefore, the authors do not stress on the specific short-term traffic forecasting sub-domain, and only a few works concerning this topic are considered.

Further away from the subject of short-term traffic forecasting, \cite{wang2020deep} revolves around spatio-temporal data mining as a general task that can be formulated in many application domains. Indeed, authors review Deep Learning models proposed for transportation and human mobility, but also take into account other unrelated topics like neuroscience and crime analysis. As a result, this survey only provides insights for some traffic forecasting solutions that benefit from spatio-temporal relationships.

Another survey on traffic prediction is available at \cite{yin2020comprehensive}, where authors summarize the state of the art on traffic prediction methods, and comment on the different Deep Learning architectures. It is the only work among those reviewed in Table \ref{tab: surveys} that performs an empirical study. This experimental setup aims for comparing performance among recent Deep Learning methods, but no further insights are given in regard to whether such performance levels are superior to those rendered by simpler learners.

Next, both \cite{tedjopurnomo2020survey} and \cite{gobezie2020machine}, provide an overview of existing Deep Learning methods for traffic flow forecasting. Future challenges for the research field are discussed in \cite{tedjopurnomo2020survey}, such as a lack of well-established benchmark datasets, the inclusion of contextual data (for instance, weather data) and the development of graph-based modeling techniques. Finally, \cite{lee2020short} conforms to a further overview of Deep Learning methods applied to short-term traffic forecasting. The authors classify published models by \emph{generation}, according to the complexity and structure of the Deep Learning technique.

After analyzing the summarized works at Table \ref{tab: surveys}, we conclude that they do not entirely provide a comprehensive, critical vision of the use of Deep Learning models for short-term traffic forecasting. Those who match the topic are restricted to an overview of the components of available Deep Learning architectures, while the remaining ones gravitate around general subjects like transportation or spatio-temporal data mining.

It is our belief that a survey should go beyond an overview of recent Deep Learning techniques, towards answering other important questions such as \emph{why?} and \emph{what for?}. Deep Learning models lead the majority of short-term traffic forecasting benchmarks, but often authors do not discuss the caveats related to their implementation. Some endemic features of Deep Learning do not comply with the requirements of traffic managers, including their computational complexity and black-box nature. Therefore, the adoption of such modeling techniques should be supported by other evidences and statements than a performance gain over other data-driven methods. Based on this rationale, this overview does not elaborate on the different Deep Learning architectures used in the literature, but instead focuses on classifying it according to alternative criteria more aligned with the questions formulated above. 

\section{Literature Review}\label{sec:review+taxonomy}

In order to acquire a thorough understanding of the current use of Deep Learning techniques for short-term traffic forecasting, in this section a taxonomy for categorizing the published works during recent years is proposed. For this purpose, previous surveys serve as a starting point towards finding the common criteria that define these categories. A literature review is performed subsequently as per the defined criteria.

\subsection{Proposed taxonomy}\label{subsec:taxonomy}

The proposed taxonomy follows two complementary strategies that recursively appear as such in the literature. The first criterion determines and characterizes the traffic forecasting problem to be solved, whereas the second criterion categorizes the Deep Learning method(s) in use for tackling it. We now describe such criteria in detail:
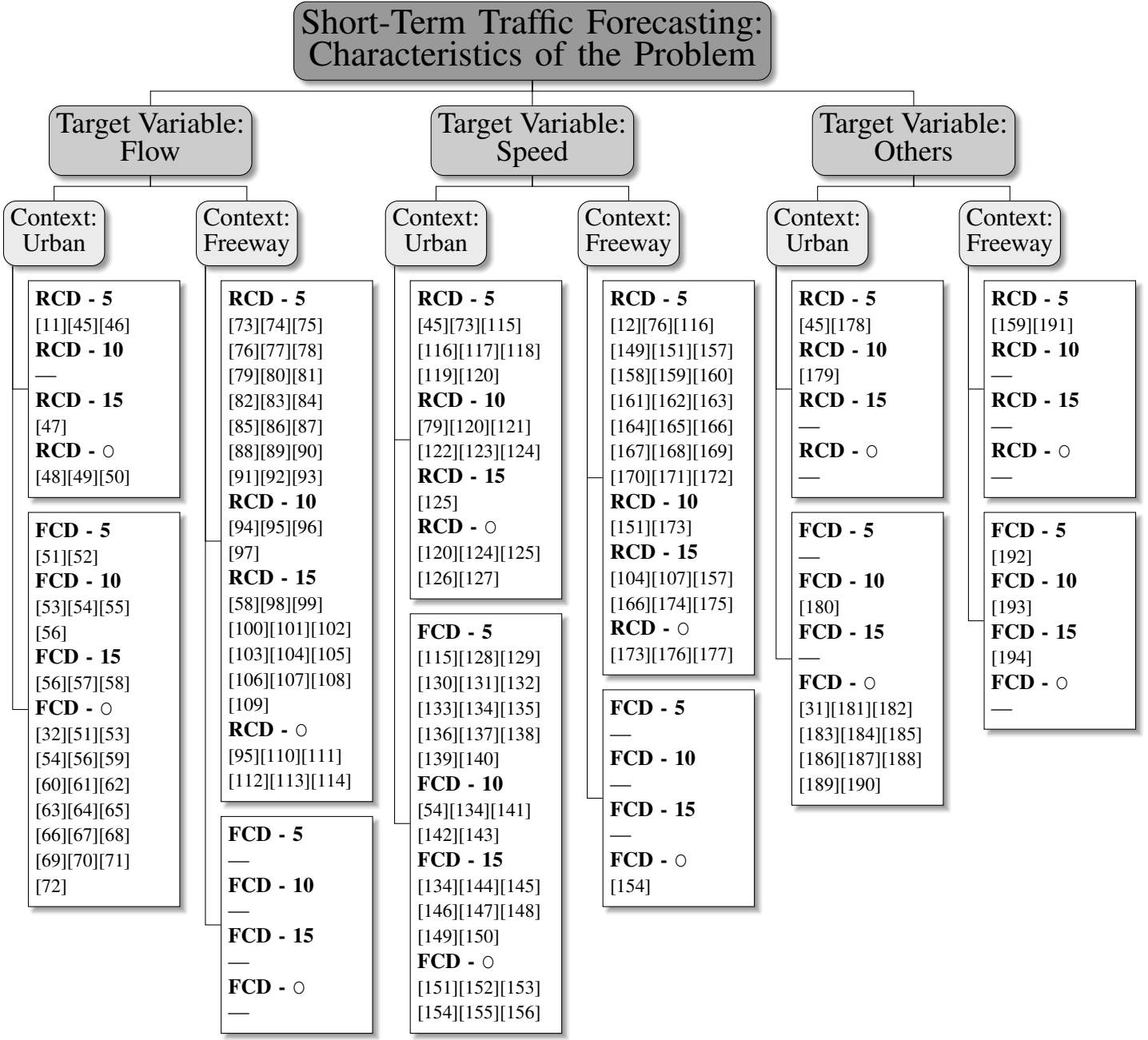
\begin{figure*}[ht!]
\begin{forest}
upper style/.style={draw,thin,blur shadow,rounded corners=6pt, fill=black!40,align=center},
lower style/.style={draw,thin,blur shadow,rounded corners=0pt,text width=6.5em,align=left},
s sep'=5pt,
forked edges,
where level<=2{%
    upper style
}{%
    lower style,
},
where level<=1{%
    parent anchor=children,
    s sep'+=1pt,
    child anchor=parent,
    if={isodd(n_children())}{%
        calign=child edge,
        calign primary child/.process={
            O+nw+n{n children}{(#1+1)/2}
        },
    }{%
        calign=edge midpoint,
    },
}{
    folder,
    grow'=0,
},
[{\LARGE{Short-Term Traffic Forecasting:}}\\{\LARGE{Characteristics of the Problem}},
    [{\Large{Target Variable:}}\\{\Large{Flow}},for tree={fill=black!20}
        [{\large{Context:}}\\{\large{Urban}},for tree={fill=black!8}
            [\textbf{RCD - 5}
            \\\small{\cite{zhao2017lstm}\cite{yi2019vds}\cite{albertengo2018short}}
            \\\textbf{RCD - 10}
            \\---
            \\\textbf{RCD - 15}
            \\\small{\cite{pamula2018impact}}
            \\\textbf{RCD - }\texttt{O}
            \\\small{\cite{bui2019big}\cite{essien2019deep}\cite{liu2017novel}}
            ,for tree={fill=white}]
            [\textbf{FCD - 5}
            \\\small{\cite{zonoozi2018periodic}\cite{sudo2017predicting}}
            \\\textbf{FCD - 10}
            \\\small{\cite{zhang2019graph}\cite{zhang2018kernel}\cite{jia2020predicting}}
            \\\small{\cite{zhang2020deep}}
            \\\textbf{FCD - 15}
            \\\small{\cite{zhang2020deep}\cite{sun2020city}\cite{huang2019diffusion}}
            \\\textbf{FCD - }\texttt{O}
            \\\small{\cite{zhang2016dnn}\cite{zonoozi2018periodic}\cite{zhang2019graph}}
            \\\small{\cite{zhang2018kernel}\cite{zhang2020deep}\cite{hassija2020traffic}}
            \\\small{\cite{ren2020hybrid}\cite{yao2019revisiting}}\small{\cite{zhang2017deep}}
            \\\small{\cite{zhang2019flow}\cite{wang2018cross}\cite{guo2019deep}}      \\\small{\cite{duan2019prediction}\cite{li2019densely}\cite{mourad2019astir}}
            \\\small{\cite{zhou2019st}\cite{chen2018exploiting}\cite{wang2018explore}}
            \\\small{\cite{duan2018improved}}
            ,for tree={fill=white}]
        ]
        [{\large{Context:}}\\{\large{Freeway}},for tree={fill=black!8}
            [\textbf{RCD - 5}
            \\\small{\cite{guo2020optimized}\cite{pholsena2020mode}}\small{\cite{dai2019deeptrend}}
            \\\small{\cite{mallick2019graph}\cite{guo2019attention}\cite{impedovo2019trafficwave}}
            \\\small{\cite{zhang2020novel}\cite{wu2018hybrid}\cite{wang2019regularized}}
            \\\small{\cite{chen2019multi}\cite{lv2014traffic}\cite{dai2017deeptrend}}
            \\\small{\cite{asadi2020spatio}\cite{asadi2019convolution}\cite{wu2018graph}}
            \\\small{\cite{fu2016using}\cite{du2017traffic}\cite{kang2017short}}
            \\\small{\cite{liu2017short}\cite{duan2016efficient}\cite{wu2016short}}
            \\\textbf{RCD - 10}
            \\\small{\cite{cai2020noise}\cite{abbas2018short}\cite{zhao2018parallel}}
            \\\small{\cite{jia2017traffic}}
            \\\textbf{RCD - 15}
            \\\small{\cite{huang2019diffusion}\cite{tian2015predicting}\cite{yang2020msae}}
            \\\small{\cite{yang2019mf}\cite{zhao2019deep}\cite{du2019lstm}}
            \\\small{\cite{manibardo2020transfer}\cite{zhang2020multitask}\cite{du2018hybrid}}
            \\\small{\cite{zhang2018traffic}\cite{elhenawy2016stretch}\cite{koesdwiady2016improving}}
            \\\small{\cite{xu2020spatial}}
            \\\textbf{RCD - }\texttt{O}
            \\\small{\cite{abbas2018short}\cite{liang2018deep}\cite{kolidakis2019road}}
            \\\small{\cite{mou2019t}\cite{zhang2018combining}\cite{yang2016optimized}}
            ,for tree={fill=white}]
            [\textbf{FCD - 5}
            \\---
            \\\textbf{FCD - 10}
            \\---
            \\\textbf{FCD - 15}
            \\---
            \\\textbf{FCD - }\texttt{O}
            \\---
            ,for tree={fill=white}]
        ]
    ]
    [{\Large{Target Variable:}}\\{\Large{Speed}},for tree={fill=black!20}
        [{\large{Context:}}\\{\large{Urban}},for tree={fill=black!8}
            [\textbf{RCD - 5}
            \\\small{\cite{yi2019vds}\cite{guo2020optimized}\cite{diao2019dynamic}}
            \\\small{\cite{george2020improved}\cite{song2017traffic}\cite{essien2020deep}}
            \\\small{\cite{liu2019traffic}\cite{shen2018research}}
            \\\textbf{RCD - 10}
            \\\small{\cite{zhang2020novel}\cite{shen2018research}\cite{piazzilstm}}
            \\\small{\cite{zhang2019link}\cite{zhang2020network}\cite{jia2016traffic}}
            \\\textbf{RCD - 15}
            \\\small{\cite{kim2018capsule}}
            \\\textbf{RCD - }\texttt{O}
            \\\small{\cite{shen2018research}\cite{jia2016traffic}\cite{kim2018capsule}}
            \\\small{\cite{ma2015long}\cite{sun2017dxnat}}
            ,for tree={fill=white}]
            [\textbf{FCD - 5}
            \\\small{\cite{diao2019dynamic}\cite{bogaerts2020graph}\cite{guo2020gps}}
            \\\small{\cite{shin2019incorporating}\cite{yu2020forecasting}\cite{cao2020hybrid}}
            \\\small{\cite{fusco2016comparative}\cite{yang2020short}\cite{han2019short}}
            \\\small{\cite{pan2019urban}\cite{sun2019traffic}\cite{zhang2019multistep}}
            \\\small{\cite{liao2018dest}\cite{wang2016traffic}}
            \\\textbf{FCD - 10}
            \\\small{\cite{zhang2018kernel}\cite{yang2020short}\cite{fu2020short}}
            \\\small{\cite{zhang2019gcgan}\cite{zhang2019trafficgan}}
            \\\textbf{FCD - 15}
            \\\small{\cite{yang2020short}\cite{zhang2019wavelet}\cite{lee2020predicting}}
            \\\small{\cite{bratsas2020comparison}\cite{liu2019deeprtp}\cite{zhang2019hybrid}}
            \\\small{\cite{zhao2019t}\cite{liao2018deep}}
            \\\textbf{FCD - }\texttt{O}
            \\\small{\cite{chen2020dynamic}\cite{zhang2018predicting}\cite{yu2017spatiotemporal}}
            \\\small{\cite{elleuch2020neural}\cite{ma2017learning}\cite{ma2020forecasting}}
            ,for tree={fill=white}]
        ]
        [{\large{Context:}}\\{\large{Freeway}},for tree={fill=black!8}
            [\textbf{RCD - 5}
            \\\small{\cite{polson2017deep}\cite{mallick2019graph}\cite{george2020improved}}
            \\\small{\cite{zhao2019t}\cite{chen2020dynamic}\cite{boquet2020variational}}
            \\\small{\cite{cui2019traffic}\cite{ryu2019intelligent}\cite{cui2020stacked}}
            \\\small{\cite{yu2017spatio}\cite{wei2019dual}\cite{zhang2019spatial}}
            \\\small{\cite{shleifer2019incrementally}\cite{manibardo2020new}\cite{yang2020evaluation}}
            \\\small{\cite{wu2019graph}\cite{yu2017deep}\cite{wang2020forecast}}
            \\\small{\cite{li2017diffusion}\cite{sun2020constructing}\cite{fandango2018towards}}
            \\\textbf{RCD - 10}
            \\\small{\cite{chen2020dynamic}\cite{jia2017rainfall}}
            \\\textbf{RCD - 15}
            \\\small{\cite{zhang2020multitask}\cite{elhenawy2016stretch}\cite{boquet2020variational}}
            \\\small{\cite{yang2020evaluation}\cite{adu2018traffic}\cite{liu2018short}}
            \\\textbf{RCD - }\texttt{O}
            \\\small{\cite{jia2017rainfall}\cite{epelbaum2017deep}\cite{he2018stann}}
            ,for tree={fill=white}]
            [\textbf{FCD - 5}
            \\---
            \\\textbf{FCD - 10}
            \\---
            \\\textbf{FCD - 15}
            \\---
            \\\textbf{FCD - }\texttt{O}
            \\\small{\cite{elleuch2020neural}}
            ,for tree={fill=white}]
        ]
    ]
    [{\Large{Target Variable:}}\\{\Large{Others}},for tree={fill=black!20}
        [{\large{Context:}}\\{\large{Urban}},for tree={fill=black!8}
            [\textbf{RCD - 5}
            \\\small{\cite{yi2019vds}\cite{cheng2018deeptransport}}
            \\\textbf{RCD - 10}
            \\\small{\cite{toncharoen2018traffic}}
            \\\textbf{RCD - 15}
            \\---
            \\\textbf{RCD - }\texttt{O}
            \\---
            ,for tree={fill=white}]
            [\textbf{FCD - 5}
            \\---
            \\\textbf{FCD - 10}
            \\\small{\cite{xu2018station}}
            \\\textbf{FCD - 15}
            \\---
            \\\textbf{FCD - }\texttt{O}
            \\\small{\cite{ma2015large}\cite{jiang2018geospatial}\cite{yao2018deep}}
            \\\small{\cite{abdollahi2020integrated}\cite{chen2020multitask}\cite{sun2020fma}}
            \\\small{\cite{ning2018predicting}\cite{rodrigues2019combining}}\small{\cite{wang2018deepstcl}}
            \\\small{\cite{saxena2019d}\cite{liao2018large}}
            ,for tree={fill=white}]
        ]
        [{\large{Context:}}\\{\large{Freeway}},for tree={fill=black!8}
            [\textbf{RCD - 5}
            \\\small{\cite{ryu2019intelligent}\cite{fouladgar2017scalable}}
            \\\textbf{RCD - 10}
            \\---
            \\\textbf{RCD - 15}
            \\---
            \\\textbf{RCD - }\texttt{O}
            \\---
            ,for tree={fill=white}]
            [\textbf{FCD - 5}
            \\\small{\cite{yi2019implementing}}
            \\\textbf{FCD - 10}
            \\\small{\cite{zhang2019deep}}
            \\\textbf{FCD - 15}
            \\\small{\cite{ran2019lstm}}
            \\\textbf{FCD - }\texttt{O}
            \\---
            ,for tree={fill=white}]
        ]
    ]
]
\end{forest}
\centering
\caption{Contributions on Deep Learning based short-term traffic forecasting reported in the literature classified according to \emph{Criterion 1}. From left to right, each branch level stands for nature of traffic measurements, traffic context, data collecting strategy, and temporal resolution in minutes. The \texttt{O} indicator refers to other less used data temporal resolutions, such as 30 minutes or 1 hour. }\label{fig:tree}
\end{figure*}

\begin{figure*}[t!]
    \centering
    \includegraphics[width=\linewidth]{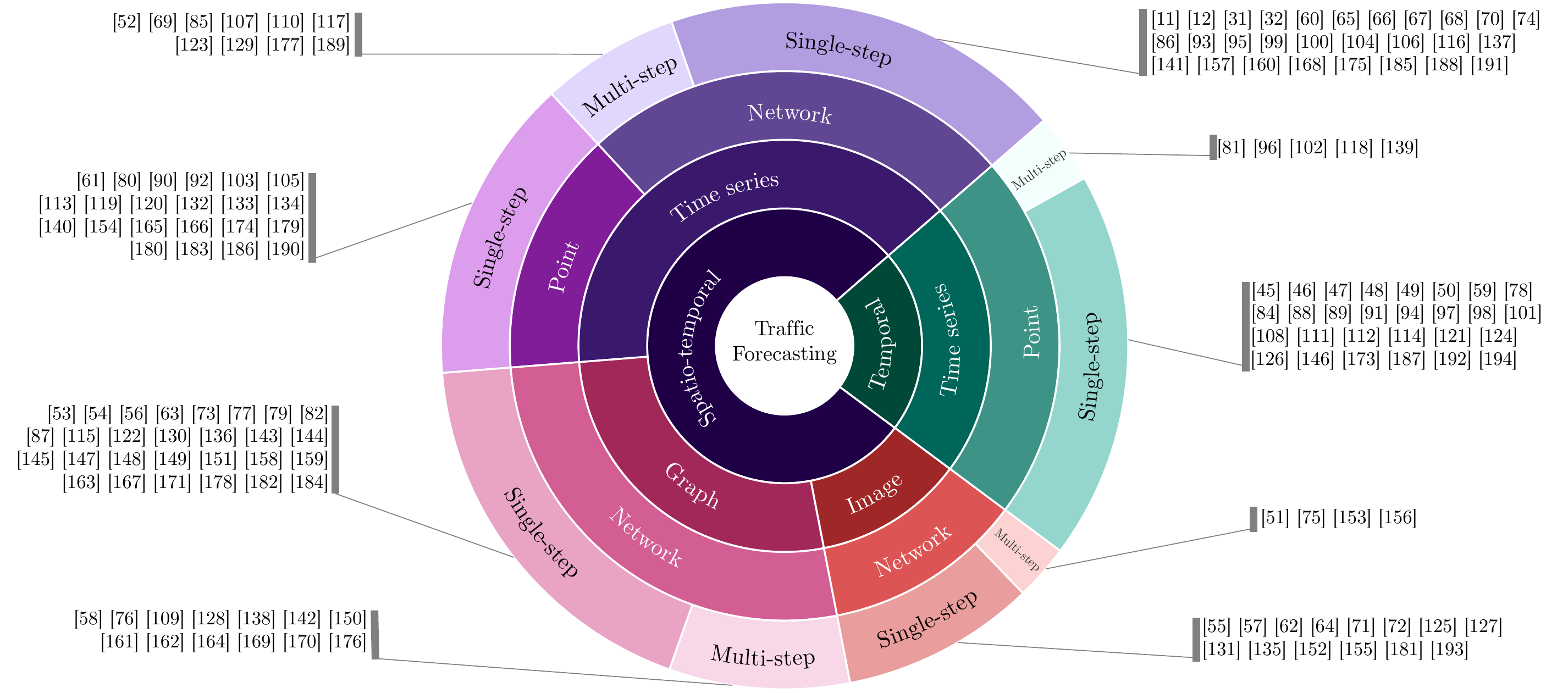}
    \caption{Deep Learning models for short-term traffic forecasting in examined works, classified according to \emph{Criterion 2}. From inside out, each ring level stands for: considered dependencies, data representation format, range of coverage, and number of steps ahead prediction.}
    \label{fig:sunburst}
\end{figure*}

\vspace{2mm}
\subsubsection*{Criterion 1. How to characterize the proposed problem}\label{subsubsec:faced problem}

Research activity of short-term traffic forecasting comprehends multiple combinations of traffic measurements, which can be combined to achieve predictions of increased quality. To illustrate the taxonomy based on the first criterion, we have constructed a tree diagram (Figure \ref{fig:tree}), which represents the patterns existing in the field. Splits' order is chosen according to their effect on the proposed problem. This way, features that yield a major discrepancy for the addressed approach are placed at higher levels of the tree, and vice versa.

Following the above guidelines, the first split is made according to the nature of traffic measurements. After reviewing the short-term traffic forecasting literature, two main strategies can be discerned: forecasting \texttt{flow}, understood as the number of vehicles that pass through the location of interest during a time interval, and \texttt{speed}, defined as the average speed over a certain time period of all vehicles that traverse the target location. Other traffic measurements are travel time, occupancy, transport user demand (e.g. for taxis or bikes) and congestion level, all grouped under the category \texttt{others}, since the number of contributions that focus on these measurements is notably lower than the previous categories.

The second split in the tree considers the traffic context: \texttt{urban} or \texttt{freeway}. The different circumstances that occur in these contexts \cite{barlovic2003traffic} generate more stable traffic patterns at highways, in contrast to urban routes, whose traffic flows are conditioned by traffic lights and signals, among other events.

The third split is set on how vehicular data are collected. Roadside sensing gathers measurements directly from road segments by using inductive loops, radar, or computer vision. On the other hand, GPS and other positioning sensing technologies allow tracking vehicle travel trajectory and speed by timestamped geolocalization measurements. These data collecting strategies are defined as Roadside Car Data (\texttt{RCD}) and Floating Car Data (\texttt{FCD}), respectively. 

The last split addresses how the collected traffic data are aggregated. Sensors can feature different sampling frequencies, from a few seconds to several minutes. Since these sampling frequencies can impose -- if high enough -- a significant variability on the traffic measurement, the collected data is usually aggregated into lower temporal resolutions. Three prediction temporal resolutions \texttt{[5,10,15]} in minutes appear to be the most commonly used ones in the reviewed literature corpus. Additionally, the \texttt{O} symbol appended at the labels of the third split refers to other less used data temporal resolutions (for instance, 30 minutes).

Before proceeding further, it is important to note that some publications may appear in multiple leaf nodes of the tree diagram. This is due to research work matching the criteria of different categories (for example, if the proposed model predicts diverse traffic measurements, or if different kind of data sources are addressed).

\vspace{2mm}
\subsubsection*{Criterion 2. How to categorize a Deep Learning technique}\label{subsubsec:followed approach}

Deep Learning architectures can be designed to adapt to diverse case studies. This design flexibility yields a heterogeneous mixture of modeling strategies. Under this premise, different features of Deep Learning methods are considered in this second criterion. A sunburst diagram (Figure \ref{fig:sunburst}) is selected to illustrate the different types of Deep Learning architectures proposed in the short-term traffic forecasting literature. The width of each angular sector is proportional to the number of research papers that fall within the category, relative to the total number of revised publications.

The most valuable information to predict the traffic state is usually that related to the target road. Previously collected data of the same road are in general good predictors of its short-term traffic profile. This statement is supported by the remarkable performance often offered by naive methods such as the historical average \cite{van2007short}, which computes the next traffic prediction value as the mean value of recent measurements at the considered point of the traffic network. On the other hand, historical information of the surrounding areas (i.e. nearby roads) and measurements of downstream and upstream points of the same road have been lately incorporated to the input of the traffic forecasting model, as they can possess interesting correlations with the traffic of the target placement \cite{cassidy1999some}. The spatio-temporal relationships between vicinity areas can provide better predictors of the traffic profile to be modeled \cite{yue2008spatiotemporal,liu2011discovering}. Those publications that feed the forecasting model exclusively with temporal data collected from the target road are categorized as \texttt{temporal}, whereas those that also resort to traffic measurements of other points in the same road network are categorized as \texttt{spatio-temporal}.

The next considered split is the format in which traffic measurements are expressed. Data related to traffic conditions are usually represented as time series, since their values are correlated through time \cite{ahmed1979analysis}. Those publications that follow a traditional time series forecasting approach are cataloged as \texttt{time series}. 

Another possible approach consists of expressing the traffic state as an \texttt{image}. The great development in Deep Learning architectures (in particular convolutional networks) has led to a revolution in the image processing field \cite{szegedy2016rethinking,he2016deep,xie2017aggregated}. In the context of traffic forecasting, the concept idea is to develop a model that predicts an image with traffic states (e.g. an image of a traffic network colored according to congestion levels). The predicted image can be transformed to express average speed, road congestion, and other traffic descriptors. Processing image representations of traffic networks allows predicting at once the traffic state at various roads of the network. 

The last considered format in this second split consists of expressing traffic data as graphs. Since traffic is restricted to road networks, it can be formulated as a \texttt{graph} modeling problem, where the structure of the road network is abstracted as a graph $\mathcal{G} = (\mathcal{V}, \mathcal{E}, \textbf{A})$ \cite{velivckovic2017graph}. In $\mathcal{G}$, $\mathcal{V}$ is a set of $N$ nodes representing road locations, whereas $\mathcal{E}$ is a set of edges representing the roads connecting such locations, and $\textbf{A} \in \mathbb{R} \textsuperscript{\textit{NxN}}$ is an adjacency matrix, in which each element $a\textsubscript{i,j}$ represents a numerical quantification of the \emph{proximity} between nodes of the network in terms of traffic flow (e.g. the reachability from one node of the graph to another, or the intensity of traffic between them). This representation of a road network and its traffic, and the use of graph embedding techniques for their input to the Deep Learning models allows providing network-wide predictions and learn from the relationships between nodes of the graph. 

Further along this second split, predictive models can be designed to forecast traffic state for one or multiple points of a traffic network. Those works that provide network-wide predictions are classified as \texttt{network}. In the case where models predict the traffic state of a single road, the research work at hand is labeled as \texttt{point}. Some studies predict different road congestion states simultaneously by using multiple models, but because the spatial coverage for each model remains to one road they are also cataloged as \texttt{point}.

The fourth considered split is the number of steps-ahead predicted by the model. For the simplest case, the model forecasts a single step-ahead point of the sequence (\texttt{single-step}), but there are models capable of predicting multiple steps ahead (\texttt{multi-step}). Another approach, known as multi-stage prediction, consists of generating a multiple steps-ahead forecasts by using a single step-ahead model, which cyclically uses as input data the recently predicted values \cite{cheng2006multistep}. As this strategy employs single step-ahead models, the corresponding contributions are classified as \texttt{single-step}.

\subsection{Understanding Deep Learning based short-term traffic forecasting literature according to the proposed taxonomy}\label{subsec:review}

Once revised works have been categorized by the proposed problem and by the chosen Deep Learning approach, an in-depth literature review is performed, in order to objectively assess the trends followed by the community in this field of research.

A first inspection of the taxonomy depicted in Figure \ref{fig:tree} reveals that the 5-minute temporal resolution positions itself as the most common in the reviewed literature. Almost half of the distinct data collections used by the reviewed papers gather data using 5 minutes sampling frequency. In addition, this trend is strengthened by the presence of Caltrans Performance Measurement System (PeMS) \cite{PeMSddbb}, which is by far the most popular traffic database, and also employs this sampling frequency. The 10 and 15 minute temporal resolution has less available original data collections, but sometimes the authors aggregate 5-minute data to obtain these resolutions, so the number of publications in this context increases slightly. Lastly, other temporal resolutions (denoted by the \texttt{O} symbol) deserve a special mention. This group merges uncommon values from 2, 3, 6, or 16 minutes to 1 or 2 hours. Some of these temporal resolutions are acquired from data collections that have been utilized only once. The 30 and 60 minutes temporal resolutions are, however, adopted in many works, usually based on FCD from taxi flow or transport user demand. Transport user demand predictions (e.g. number of bikes expected to be rented during a time interval) usually employs low temporal resolutions, as these rates suffice for capturing the collective behavior of the population.

When focusing on traffic flow forecasting models, there is a clear tendency towards using RCD from freeways. The high cost of roadside sensors makes them to be typically deployed on critical road sections such as freeways, so there are more data sources of this kind than from urban arterials. However, since RCD is highly biased by the deployment location, its potential is limited when developing general-purpose traffic forecasting models. Interestingly, there are not FCD based reviewed works that forecast freeway flow. FCD that captures flow measurements is mainly obtained from taxis and logistics services, or from passengers carrying cell phones in the vehicle. In the case of urban flow prediction, there are several published works, yet the majority of them are conducted over taxi or bike floating data. Since this sensing technique only captures a fraction of the circulating vehicles, FCD is usually utilized to predict flow values of certain vehicles type, and is hence not suitable for \emph{general} flow forecasting problems. Research contributions are more balanced towards traffic speed prediction, covering all data type and granularity combinations, except for FCD at freeways, where only one work has been found \cite{elleuch2020neural}. PeMS and Los Angeles County highway dataset (METR-LA) \cite{jagadish2014big} are the preferred option when looking for freeway speed RCD. For speed prediction task, FCD provide reliable measurements since the average speed of the sensed vehicles (even though it is only a part of the vehicle fleet) can be considered as the average circulation speed on the road for an specific time interval.

Lastly, the \texttt{others} label blend together a mixture of works that predict traffic congestion \cite{ma2015large,cheng2018deeptransport,toncharoen2018traffic,fouladgar2017scalable,zhang2019deep}, expected travel time \cite{ryu2019intelligent,abdollahi2020integrated,ran2019lstm}, occupancy \cite{yi2019vds} and traffic performance index \cite{yi2019implementing}. A special mention must be made to those works which predict service demand, understood as the number of vehicles necessary to cover a passenger demand. In this context, taxi demand is the most covered scope, probably due to the high data availability \cite{jiang2018geospatial,yao2018deep,chen2020multitask,rodrigues2019combining,wang2018deepstcl,liao2018large}. There are also works focused on sharing bike demand \cite{xu2018station, saxena2019d}. In either case, the \texttt{others} label covers different combinations of data types and temporal resolutions, so there is not a clear trend in this subset of contributions.

When the focus is placed on the employed methodology, Figure \ref{fig:sunburst} unveils a clear increase of published studies that combine spatial and temporal information over recent years \cite{ermagun2018spatiotemporal}. There are three times more works of this nature compared to those based only on temporal information. For a publication to be classified as \texttt{temporal}, the presented study can only take advantage of the knowledge from historical records at the point for which a predictions is issued. Therefore, the input format can only be classified as \texttt{time series}, since \texttt{image} and \texttt{graph} data representations always express information from multiple points of a traffic network. In turn, if we combine the number of publications based on temporal information with those based on spatio-temporal information, it can be seen that more than half of the works formulate the input data as time series, which is the basic format to express traffic state.

As stated in the work of Ermagun et al. \cite{ermagun2018spatiotemporal}, the number of works based on graph theory \cite{bollobas2013modern} has increased notably in recent years. Describing a traffic network as a graph adds spatio-temporal relational information between the different places where traffic state prediction is required, providing network-wide forecasts. For the remaining input formats, traffic representation as an image is the least chosen option, with about an eighth part of reviewed works. Some of these studies generate images from time series transformations of different points of the network expressed as matrices. Since the model is fed with images, even if they are a representation of multiple time series, these publications are classified as \texttt{image}. 
Graph based, image based, together with some time series based model works, represent more than half of revised publications dealing with network-wide coverage solutions. While these studies usually concentrate on performing simultaneous predictions for multiple traffic network points, publications classified as \texttt{point} often put their effort on other specific issues like traffic signal processing \cite{kolidakis2019road, cao2020hybrid}, the exploration of new data sources \cite{hassija2020traffic,essien2020deep}, the improvement of performance under particular situations \cite{manibardo2020transfer,manibardo2020new} or missing data \cite{pamula2018impact,cui2020stacked}. 

Finally, single-step models represent the majority of existing publications, as is in general an easier modeling task when compared to multi-step prediction. However, there is a surprisingly high amount of contributions (17.6\%) that provide network-wide multi-step prediction, considering the difficulty of predicting multiple steps-ahead of traffic state values for different locations simultaneously.

\section{Critical Analysis}\label{sec:criticalanalysis}
A critical look to the preceding literature review raises some questions about the suitability of Deep Learning techniques for the task of short-term traffic forecasting: is it always the best choice? In this section, the main aspects of this consideration are assessed trying to answer to eight questions, and examined towards opening a debate:
\begin{enumerate}[label=\emph{\Alph*.},leftmargin=*]
    \item When is a forecast considered long-term?
    \item Are traffic datasets correctly selected?
    \item Can Deep Learning models be trained with scarce data?
    \item Does the use of contextual data yield any benefit?
    \item Is data representation an objective or a circumstance?
    \item Is automatic feature extraction interesting for traffic data?
    \item What possibilities does data fusion offer?
    \item Are comparison studies well designed?
\end{enumerate}

\subsection{When is a forecast considered long-term?}

The use of Deep Learning techniques for traffic forecasting is relatively recent \cite{do2019survey}. However, the frontier between short and long-term predictions seems to remain largely ambiguous for many authors, thus jeopardizing the identification of Deep Learning models devised to tackle one or the other problem. This lack of consensus hinders the proper selection of modeling counterparts in benchmarks arising in the newer studies, often featuring an assorted mixture of short- and long-term approaches.

Authors of some related works establish the distinction between short- and long-term forecasting in terms of the prediction horizon, claiming that a prediction further than one hour ahead should be considered as long-term. This is by all means an unreliable consideration since, for a model where the time between consecutively arriving samples is one hour, a one-hour-ahead prediction problem would translate to a one-step-ahead forecasting task. There are other shared interpretations by which short-term forecasting is assumed to cover only the very first time steps (usually no more than five steps) disregarding the temporal resolution of the time series at hand. However, for a fixed temporal resolution, models can be prepared to directly output one particular forecasting horizon (e.g. twelve-step-ahead). This case would entail some authors to classify it as long-term, while others would claim that it is short-term prediction, as the model is trained to forecast only that specific time step.

In our best attempt at homogenizing the meaning of these concepts, we herein clarify the applicability of both approaches. Short-term predictions allow travelers to select among more quick and efficient routes, by avoiding bottlenecks. Likewise, local authorities can quickly respond and hopefully circumvent traffic congestion. They are, therefore, \emph{operational} models \cite{naess2015traffic}, which predictions are restricted to delimited geographical areas, since the interactions of the surroundings affect the traffic itself. On the other hand, long-term estimations allow traffic managers to prepare and implement \emph{strategic} measures in case of predictable eventualities, such as sports events, weather conditions, road pricing, or general strikes \cite{lamboley199724}. The management of large areas (i.e. city-wide) may improve, for example, the design of road side infrastructure \cite{jha2016comparative}, eventually leading to more fluent traffic.

Based on this rationale, short-term models are usually built based on recent past observations of the target road and its vicinity to estimate their immediate subsequent ones. Here is where the distinction between approaches can be made: the model construction methodology. Long-term traffic estimation models seek different traffic patterns (e.g. typical daily traffic profiles), and decide which of these patterns suits best the traffic behavior of the selected road for the date under choice \cite{lana2019question}. The chosen pattern among all those elicited by the model becomes the prediction for the entire interval. Therefore, long-term estimation is, in general, less accurate and prone to larger errors in the presence of unexpected circumstances or when the selected output traffic pattern is inaccurate. By contrast, they provide a general idea of the expected behavior that can be used by traffic managers to decide strategic measures. Short-term forecasting models, on the other hand, issue their predictions by learning from recent past observations, obtaining more reliable forecasts as the model has access to better predictors for the target variable.

\subsection{Are traffic datasets correctly selected?}

Presented literature review unveils another issue: the majority of publications select only one data source or multiple of the same scope (for instance, traffic collected in highways or urban areas, but not from both in the same study). This trend is observable by placing attention on duplicated citations at different leaves of the tree diagram in Figure \ref{fig:tree}. Benchmarks comprising datasets of different characteristics is a good practice that should be widely adopted for assessing the performance of newly proposed Deep Learning models. As addressed in Section \ref{subsec:taxonomy}, there are some characteristics of a traffic forecasting problem that can appreciably affect the model performance, namely data source type, data source context, predicted variable.

From the perspective of the data source type, RCD is an integrated count of any transportation vehicle that passes through the sensor location, while FCD is usually collected by vehicle types like taxis, buses, trucks, or bikes. The different way in which these two data types are gathered can impact severely on the time series behavior, leading to mismatches in the performance comparison. Besides the data type, the data collecting context is also relevant. Urban traffic is regulated by road signs and light traffics, leading to a particular driving behavior with higher data dispersion. On the other hand, freeway traffic forecasting is an easier task when compared to urban, since traffic profiles are usually more stable in the absence of traffic signs, pedestrians and other urban circumstances. Lastly, the different predicted variables (flow, speed, travel time) can express traffic congestion states but have different profiles and behaviors. Traffic speed measurements conform to a stable signal over time that exhibits scarce yet deep valleys when a traffic bottleneck occurs. By contrast, traffic flow measurements often show different kinds of daily patterns, where the difficulty resides in predicting sudden spikes.

To sum up, a Deep Learning architecture providing good performance results for a certain traffic data source could fail to generalize nicely to other traffic data sources with different characteristics. This behavior can be detected after testing a proposed Deep Learning method, along with a mixture of data-driven algorithms, to a collection of traffic data sources with varying characteristics. Otherwise, the novelty of the proposed model should be circumscribed to the characteristics of the traffic data source(s) over which it has been tested, rather than claiming for a superior model for traffic forecasting in the wide sense.

\subsection{Can Deep Learning models be trained with scarce data?}

The ITS community has leaned towards Deep Learning based on the premise that these techniques can extract knowledge from unprocessed data more effectively than shallow learning methods. This mindset might be mistaken, as shallow learning models are advantageous in scarce data scenarios.

The main reason for it is that shallow learning models often require fewer parameters to be fit, leading to faster and less computationally demanding training processes, but also to less complex models. Since Deep Learning architectures have a potentially large number of trainable parameters, larger datasets are needed to prevent algorithms to learn the detail and noise in the training data (\emph{overfitting}), to the extent that, unless properly counteracted, it negatively impacts on the performance of the model in real-life scenarios. 

Therefore, in the context of scarce data, shallow learning methods may overcome the performance of Deep Learning models whenever the validation and test stages are designed and carried out correctly. Some of the works analyzed in our literature study consider very small periods of traffic data for training and testing. It could be thought that the results of these works are biased, since one could intuitively expect that the traffic behavior changes between months, weekdays, and daily hours \cite{ivan2002estimating}. 

As an example, if a forecasting model is trained over February data, and tested over measurements collected in March, both winter months have similar traffic behavior. This issue with a limited training data context is precisely the case where Deep Learning is prone to overfitting, leading to a higher yet biased performance on a test set. After enough training epochs, the model is good at the exposed scenario: forecasting traffic at winter, non-vacation months. This means that this Deep Learning model can be proficient forecasting in these highly specific circumstances, but will probably have trouble to generalize to other scenarios, rendering it useless. Since shallow learning methods usually have less trainable parameters, they can potentially outperform Deep Learning models in this scarce training data scenario, due to a less overfitting over data distribution.

In order to avoid overfitting of the model, the training samples - trainable parameters ratio should be maintained high, and the more trainable parameters of a model, the more training data should be required \cite{vabalas2019machine}. If this availability does not hold, the results of Deep Learning modeling experiments can be excellent due to overfitting, and be far from the good generalization properties sought for realizable traffic forecasting, which can lead to inconclusive insights.

\subsection{Does the use of contextual data yield any benefit?}

The performance of predictive models can be improved with information that does not directly express the road traffic state. We refer to it as \emph{contextual data}, since this data indicates temporal, meteorological, social, or other circumstances that can indirectly influence traffic profile. Calendar information \cite{laverty1998cyclical}, usually discretized as $[weekday, saturday, sunday]$, is commonly used as an additional source of knowledge \cite{ren2020hybrid, yang2019mf, zhang2018predicting, jiang2018geospatial}, supported by the intuition that traffic profile varies between workdays and weekends \cite{weijermars2004daily}. Another option is to provide the interval of the day, ensuring that the learning algorithm is able to correlate the temporal instant with traffic peaks \cite{song2017traffic, bogaerts2020graph,guo2020gps,fouladgar2017scalable}. Weather has also been shown to affect drivers' behavior, eventually having an impact in the overall traffic \cite{maze2006whether}. Precipitations, wind, fog, and extreme temperatures are considered as model inputs in many traffic forecasting publications, intended to help predicting unusual traffic profiles \cite{essien2019deep,ren2020hybrid, yang2019mf,ryu2019intelligent}. In this line, air pollution can be used as a congestion predictor, based on the idea that certain pollution gases (for instance, \textit{\ch{CO}, \ch{CO2}, and \ch{NOx}}) are expelled by exhaust systems. Therefore, air pollution should increase during traffic congestion and high occupancy periods, so models can benefit from this relationship \cite{zhang2020deep,awan2020improving}. Lastly, other events like manifestations, sports games, or accidents can be fed to forecasting models in order to identify uncommon traffic profiles \cite{ zhang2016dnn,zhang2017deep, yao2018deep, fouladgar2017scalable}. 
In what regards to Deep Learning models, the inclusion of previously described contextual data does not differ from its implementation with other Machine Learning models. These contextual data can be expressed as time series (e.g. temperature or air pollution), or as a discrete sequence of finite values (for instance, calendar information or timestamp). Just by increasing input dimensionality, both Deep and Machine Learning models can append new sources of knowledge towards enhancing forecasting performance. However, within the bounds of network-wide traffic predictions, Deep Learning architectures stand out in the use of contextual data. The model can be fed with dedicated contextual data for each node of the traffic network, such as accidents or road cuts. This inherent capability of Deep Learning allows flexible solutions where contextual data serve as input only by demand at specific points of the neural network, avoiding output prediction noise due to high dimensionality inputs.

\subsection{Is data representation an objective or a circumstance?}

As previously explained, short-term forecasting models are usually built upon recent past traffic state observations. The most common option, as it can be observed in Figure \ref{fig:sunburst}, is to express traffic measurements as a vector for single road state prediction, or as a matrix for multiple-point prediction. Some researchers transform traffic time series into images, and estimate the images that best represent the network behavior at the time horizon for which the prediction is issued. Other authors instead design graph representations of the traffic network, aiming to learn from the spatial relationships between nodes.

However, the choice of data representation format does not always respond to a practical consideration. Sometimes, the actual contribution of a published work is to effectively adapt traffic forecasting tasks to image-based Deep Learning architectures. The method with which the traffic information is transformed into an image is the claimed cornerstone of the proposed method. However, this traffic representation does not add any valuable knowledge to the field, as it is just another way of expressing a time series. When describing a network as a matrix, its structure predetermines the connections between the analyzed roads that a Deep Learning architecture is able to model. Convolutional filters (which are commonly used for image processing) usually look for adjacent values to discover interesting high-dimensional features, so the same information arranged differently can produce contrasting performance results. Moreover, the complexity of an actual road network can hardly be represented only by the nodes that have sensors on them (which are the ones considered for any data-based study). Thus, the \textit{picture} that represents the road network is distorted with regard to the actual road network. For a convolutional filter, the adjacency of two pixels has a particular meaning in the way they are processed, but this adjacency can have very different meanings within a network in terms of real adjacency. Hence, the claimed "spatial" awareness that this kind of methods provide must be handled with caution. Anyhow, traffic forecasting as an image can be interesting when the inputs are indeed images, for instance, screenshots from navigation services, satellite imagery, or other similar sources, as this is its original data format. 

On the other hand, graph theory suits better for network representations, by providing node relationships (both directed and non-directed variants \cite{bollobas2013modern}), which are indeed supplementary information. The underlying structure of traffic data conforms a non-Euclidean space, as a traffic network can not be modeled in a $n-$dimensional linear space without losing information (for instance, direction of the edges or values associated to nodes) \cite{bronstein2017geometric}. It is for this reason that graph representations are best suited for network-wide forecasting models, where topological information of the traffic network can be fully exploited by the model. In the case where graph modeling is not an option (e.g. unclear node assignment), time series arranged as a matrix provides a flexible and straightforward format. 

\subsection{Is automatic feature extraction interesting for traffic data?}

As previously stated in Section \ref{subsec:deep}, the most recognized capability of Deep Learning models is their ability to learn hierarchical data representations autonomously, overriding the need for handcrafting features from traffic data. As per many related studies, it is often argued that any non-Deep Learning based traffic prediction model potentially achieves a lower performance due to the fact that Deep Learning is able to model long-term dependencies in data (as opposed to handcrafted features). However, this point of view can be debatable.

Feature engineering is a difficult task that requires time, effort and domain knowledge from researchers. Nonetheless, the problem is that the predictive power of the produced features directly conditions the performance of prediction models. When input data is not self-descriptive and genuine features are not available, Deep Learning may outperform shallow learning due to its capability to learn from raw data. Nevertheless, traffic data used as inputs for traffic forecasting directly express traffic state. As an example, when the average speed of the road is available, the speed value determines if drivers are facing a \emph{free-flow} traffic state or different severity levels of bottlenecks. The model only needs to interpret these values to output a proper prediction, and probably will not need any additional features.

Traffic observations can indeed be processed to obtain more complex and specific indicators \cite{cassidy1999some,bertini2005empirical}, but models are often trained upon raw traffic data. Thus, it could be said that the feature values automatically extracted by Deep Learning architectures in recurrent networks are in fact, the extraction of long-term patterns, since short-term dependencies can be modeled by a multi-layer perceptron or other basic models. Furthermore, given the nature of the data handled in traffic forecasting, in many occasions the expert knows the recurrence patterns in advance, which makes the feature learning capability of Deep Learning less relevant for the prediction task.

In summary, automated feature extraction is a powerful feature of Deep Learning, but in the context of traffic forecasting it could not be a deciding factor for selecting this modeling approach against other data-driven methods. 

\subsection{What possibilities does data fusion offer?}

In addition to traffic recordings, other types of data sources may improve the prediction accuracy of traffic forecasting models. Beyond the feature mapping capacity of Deep Learning methods, a motivational driver for using these techniques should be its capability for in-model data fusion.

Data fusion is defined as the capacity for automatically or semi-automatically transform information from different sources into a representation of the modeled process \cite{meng2020survey}. In this context, there are some data abstractions that can not be processed by shallow learning methods. For instance, graph theory is able to model traffic network topology, and therefore the relationships between neighboring interconnected roads. Researchers take advantage of this representation via graph embedding layers to enhance the overall prediction performance of the model, as it can learn the traffic stream direction directly based on how the nodes of the graph are connected \cite{guo2020optimized,lee2020predicting, wang2020forecast}. Another example is text data, which is often asynchronously generated. There are some works that use Twitter messages \cite{essien2020deep} or queries issued for the same destination in a navigation service as congestion predictors \cite{liao2018deep}. Images are also data representations that can be processed by Deep Learning architectures. Some studies arrange snapshots of network-wide traffic congestion maps as a time series, and resort to Deep Learning architectures for motion prediction to estimate the future trajectory of objects \cite{jia2020predicting,zhang2019deep}. Other works convert traffic speed time series from multiple points of a traffic network into a heatmap, where color expresses the speed value \cite{kim2018capsule,ma2017learning}. All these examples illustrate the way in which data fusion capabilities can be used to take advantage of the Deep Learning methods potential.

Finally, complex neural architectures can assimilate on-demand specific data sources like weather or air pollution, by directly inserting these features at specific layers (generally after convolutional and recurrent layers, as these data do not need feature mapping). The model would use this information only when needed (e.g. during a special event like a football match), disabling these inputs during normal operation, to reduce model output noise. It does not seem that the traffic forecasting research community has taken advantage of this capability, which could be considered even more interesting for this particular field that its feature extraction competence.

\subsection{Are comparison studies well designed?}

The heterogeneity of methodological procedures for comparing traffic forecasting models is also visible in the literature review. For the comparison to be useful for the community, methodologically principled comparisons should be performed. Otherwise, the reported results in upcoming literature might be misleading, and disguise the real performance of novel traffic forecasting methods. For instance, some works compare their proposed model to simpler Deep Learning architectures. Instead, other contributions choose a mixture of naive, statistical, and Deep Learning models, but miss to include any kind of shallow learning method in the comparison. This variability of comparison methodologies make such studies inconclusive. In order to provide verifiable evidence of the performance improvement achieved by the proposed model, several baselines combined with state-of-the-art methods should be analyzed and compared to each other.

Starting with those methods without complexity, a few revised papers include a naive model as a baseline. These low-complexity straightforward methods have two main representatives: latest value (LV) (also referred to as \emph{persistence}) and historical average (HA) \cite{van2007short}. Since LV uses the most recently recorded traffic value as its prediction, no further calculation is required. On the other hand, HA consists of averaging past traffic data of the same interval of the day and weekday to produce the forecasting value of perform some sort of rolling average over the latter available values. This way, HA requires past sample values for computing the mean for every new prediction. In fact, HA should take into account the patterns that the expert knows in advance (for example, daily and night traffic patterns). Due to their low computational effort, at least one naive method should be considered in the comparison study, as they establish the lowest performance expected to be surpassed by a more elaborated model. If a novel forecasting method performs slightly better, equal or even worse than naive methods, the complexity introduced during training would render this method irrelevant to solve such forecasting task. Therefore, these naive methods allow assessing the balance between the complexity of the proposed model and its achieved performance gap.

Some works revised in the literature analysis compare a novel Deep architecture against different statistical methods (for instance, an ARIMA model). These methods can be set as a performance baseline, but their parameter tuning should be fully guaranteed to ensure that the statistical model is properly fit to the traffic data. According to \cite{karlaftis2011statistical}, the comparison between statistical and neural network models is unfair, as complex nonlinear models are compared to linear statistical models, drawing attention to performance metrics. Unfortunately, our literature study confirms that this malpractice still can be found in recent research. The aforementioned naive methods also provide lower bounds for the performance of traffic forecasting models. As opposed to statistical methods, they do not have adjustable parameters, so naive methods can provide a more reliable baseline for distinct traffic forecasting scenarios. Furthermore, the community could be overlooking other benefits carried by statistical methods, such as their ability to provide insights on the data and its structure. 

Simple neural architectures should not be the only ones chosen for comparing newer Deep Learning proposals (for example, stacked auto-encoders). The recent literature should be revised to elaborate comprehensive comparison studies, not only with basic Deep Learning architectures that presumably will perform worse than the proposed method, but also with the latest novel architectures, especially for spatial-temporal modeling (e.g. graph convolutional networks). 

Finally, it should be highlighted that almost none of the revised works provides complexity measures for the models under comparison. Complexity is usually quantified by the number of internal parameters to be fit. Another well-established metric is the raw training time, always determined under identical conditions (i.e. same train data collection, computing resource and software framework). After building a performance benchmark, adding complexity measures should be mandatory for the sake of fairness in comparisons. With each passing year, it becomes more difficult to overcome the performance of previous proposals, narrowing the room for improvement between the latter and the emerging architectures. In this context, these measurements provide an objective tool to judge whether the complexity introduced in the novel traffic forecasting method compensates for the performance gain over the last dominating technique. Only in this way it can be verified whether the proposed model yields an effective and efficient improvement for traffic forecasting.

\section{Case Study}\label{sec:casestudy}

From our previous analysis we have concluded that the application of Deep Learning methods to short-term traffic forecasting has been, to a point, questionable. In some cases, authors do not justify the high computational complexity inherent to their proposed method, nor do they compare it to less complex modeling alternatives. In turn, the  configuration of the comparison studies and the lack of depth in the discussion and analysis of the obtained results do not often clarify whether newly proposed methods outperform the state of the art at the time of their publication.

This section describes a case study, which serves as an informed assessment of the effects of all the particularities of the Deep Learning methods previously described. To this end, the effectiveness of these techniques when predicting short-term traffic measurements is verified and compared to modeling techniques with less computational complexity.

\subsection{Experimental setup}

A traffic forecasting case study has been designed with the aim to showcase all the details and obstacles that come along with a Deep Learning comparison study. The critical literature analysis has demonstrated that Deep Learning is a suitable option for modeling spatio-temporal relationships (whenever enough data granularity is available for such relationships to be of predictive value for the traffic state to be predicted), or to map data that are not available as time series. Those solutions that address the problem as a general time series forecasting problem disregarding the nature of the time series (i.e. the same techniques would be applied for medical or stock market time series) are defined as \emph{conventional} time series approaches. They only employ past traffic measurements as the input to the model since these features are good descriptors of the future traffic state. Deep Learning can predict graph or image representations that express network-wide traffic areas, but for conventional time series forecasting, the choice of such complex and computational consuming techniques must be solidly justified.

To shed light on this matter, a case study is designed where the goal is to resolve a traffic time series forecasting problem. According to the proposed taxonomy, traffic forecasting setups can differ in the nature of traffic measurements, the area under scope, the sensing technique, and the way data are aggregated. Although the intention is to emulate all possible cases, the number of possible setup combinations is high, so a representative subset of problems has been selected.

As shown in Figure \ref{fig:tree}, traffic flow and speed forecasting are the traffic measurements mostly addressed by the works revised in our literature study. While both time series are related by the fundamental diagram of traffic flow \cite{geroliminis2011properties}, predicting speed is in general an easier task since, for most of the time, traffic circulates at the speed limit of the road (\emph{free-flow}). It is, therefore, a more stable -- hence, predictable -- signal over time. However, traffic flow has a wider dynamic value range, and in general undergoes multiple variations throughout the day. Likewise, drivers introduce different behaviors in cities \cite{adamidis2020effects}. Urban trips are exposed to a manifold of factors such as roundabouts, pedestrian crossings or traffic lights. These aspects make data noisier and hence harder to predict. In contrast, highway traffic is not affected by such factors, so forecasting freeway traffic is in general much easier. 

Based on the above reasons, at least four datasets should be needed to cover all possible combinations of flow and speed forecasting over urban and highway areas. Table \ref{tab: data sources} summarizes the attributes of each selected data source according to the taxonomy defined in Section \ref{subsec:taxonomy}. All data sources gather traffic information by using roadside sensors. To the best of our knowledge, no public FCD data source covers one complete year of data, which is a requirement to gauge the perform of the model throughout all seasons of the year. The temporal resolution is kept to the original value provided by the data repository. 
\begin{table}[!ht]
\caption{\label{tab: data sources}Selected data sources and their characteristics.}
\resizebox{\columnwidth}{!}{
\begin{tabular}{lccccc} 
 \toprule
 \multirow{1}{*}{\textbf{Location}}  &
 \multicolumn{1}{c}{\textbf{Traffic variable}}  &
 \multicolumn{1}{c}{\textbf{Scope}}   &
 \multicolumn{1}{c}{\textbf{Sensor}}   &
 \multicolumn{1}{c}{\textbf{Time resolution}} &
 \multicolumn{1}{c}{\textbf{Year}} \\
 \cmidrule(lr){1-6}
 Madrid \cite{Madridddbb} & Flow  & Urban  & RCD   & 15 min &  2018\\
 California \cite{PeMSddbb} & Flow  & Freeway  & RCD   & 5 min & 2017 \\
 New York \cite{NYCddbb} & Speed  & Urban  & RCD   & 5 min & 2016 \\
 Seattle \cite{Seattleddbb} & Speed  & Freeway  & RCD   & 5 min & 2015\\
 \bottomrule
\end{tabular}
}
\end{table}

The forecasting problem is formulated as a regression task, where the previous measurements of each target road collected at times $\{t-4, \dots ,t\}$ are used as features to predict the traffic measurement at the same location and time $t+h$. Four prediction horizons $h\in\{1,2,3,4\}$ are considered, so that a separate single-step prediction model is trained for each $h$ value and target location.

\begin{figure}[h]
\includegraphics[width=\columnwidth]{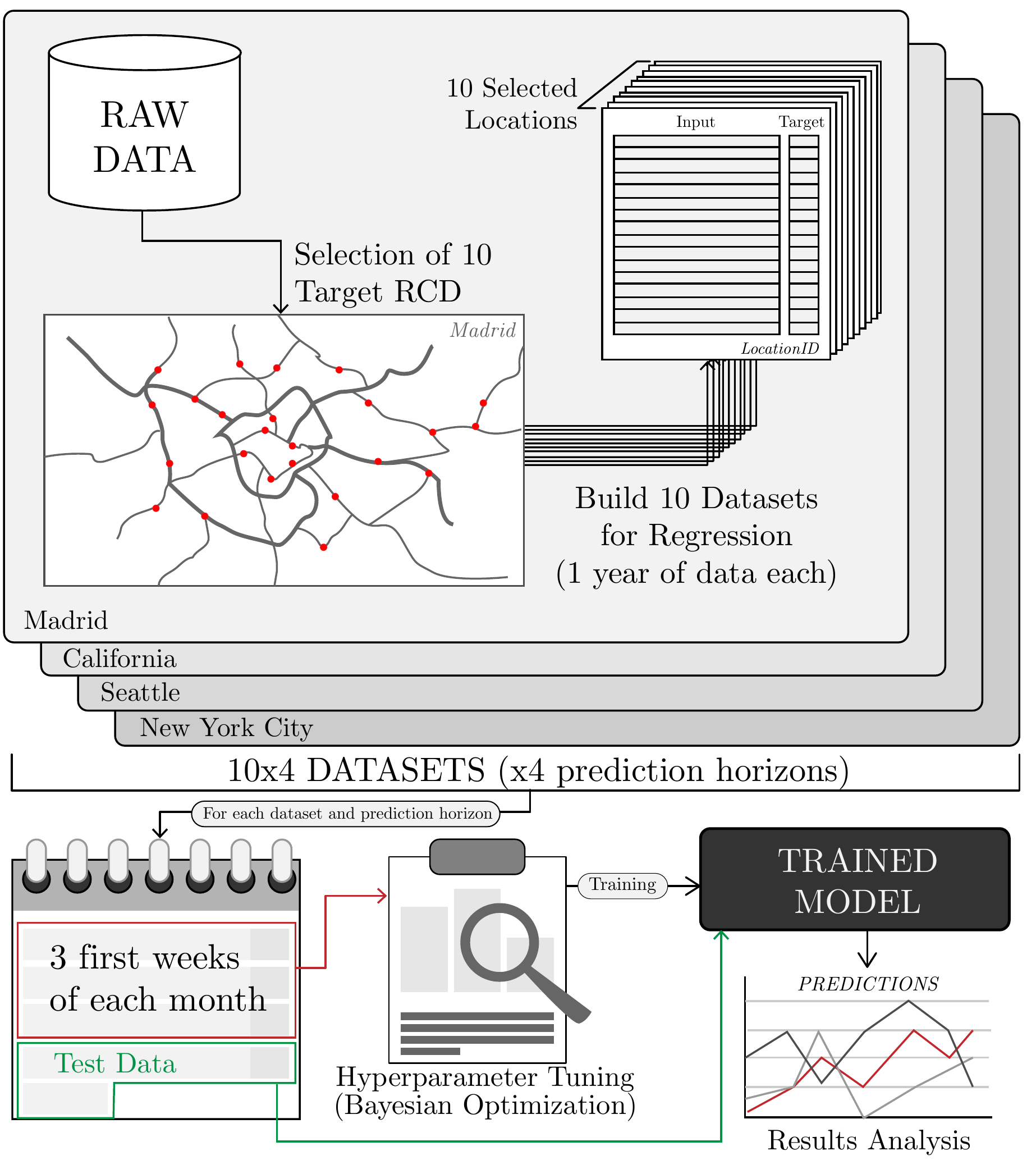}
\centering
\caption{Experimental setup used in our case study. After building regression datasets for every target location, training and testing data is reserved for every month along the year. Cross-validation provides measures to select the best hyper-parameter configuration for every model in the benchmark via Bayesian optimization. Finally, the optimized model learns from all available training data, and predictions are generated for all testing data.}
\label{fig:expsetup}
\end{figure}

Figure \ref{fig:expsetup} describes the proposed experimental setup. For each traffic data source, 10 points of the road network are selected, always choosing locations that offer diverse traffic profiles. Then, a regression dataset for each target placement is built, covering data of one year. The first three weeks of every month are used for model training, whereas the remaining days are kept for testing. This split criterion allows verifying whether models are capable to learn traffic profiles that vary between seasons and vacations days. 

In order to find the best hyper-parameter values for each regression model, three-fold cross-validation is performed: two weeks of every month are used for training, and the remaining ones of the reserved training data are used for validation. The average of the three validation scores (one per every partition) is used as the objective function of a Bayesian optimizer \cite{bergstra2013hyperopt}, which searches for the best hyper-parameter configuration efficiently based on the aforementioned objective function. After evaluating 30 possible configurations for each model, the best hyper-parameter configuration is set on the model at hand, which is trained over all training data. Once trained, model performance scores are computed over the data held for testing. This process reduces the chances to have a bias in the comparisons later discussed due to a bad hyper-parameter configuration of the models.

The purpose of the case study is to identify the model that best predicts the traffic signal for each of the prediction horizons. To this end, we compute the $R^2$ score over the testing data to measure the quality of predictions between real and predicted traffic measurements. 
This score is given by:
\begin{equation} \label{eq:r2}
R^{2} \doteq 1 - \frac{\sum_{t\in \mathcal{T}_{test}}\left(o_t - \widehat o_t\right)^2}{\sum_{t\in \mathcal{T}_{test}} \left(o_t - \bar{o_t}\right)^2},
\end{equation}
where $\mathcal{T}_{test}$ denotes the set of time slots belonging to the test partition of the dataset at hand, $o_t$ denotes the real observed value at test time $t$, $\bar{o_t}$ its average, and $\widehat{o_t}$ the predicted one.

The forecasting methods that will compose the benchmark are selected from the most commonly used algorithms and architectures in the state of the art. Statistical methods are not included in this case study, since the naive LV method already provides a performance baseline that suggests interesting insights in the experimentation. Inspired by revised works, a categorized list of learning methods is presented:
\begin{labeling}{Ensemble}
\item [\textit{Naive}] Latest Value [\texttt{LV}].

\item [\smash{\stackunder{\textit{Shallow}}{\textit{Learning}}}] Least-squares Linear Regression [\texttt{LR}], k Nearest Neighbors [\texttt{KNN}], Decision Tree [\texttt{DTR}], Extreme Learning Machine [\texttt{ETR}] and $\varepsilon$-Support Vector Machine [\texttt{SVR}].

\item [\smash{\stackunder{\textit{Ensemble}}{\textit{Learning}}}] Adaboost [\texttt{ADA}], Random Forest [\texttt{RFR}], Extremely Randomized Trees [\texttt{ETR}], Gradient Boosting [\texttt{GBR}] and Extreme Gradient Boosting [\texttt{XGBR}].

\item [\smash{\stackunder{\textit{Deep}}{\textit{Learning}}}] Feed Forward Neural Network [\texttt{FNN}], Convolutional Neural Network [\texttt{CNN}], Recurrent Neural Network based on LSTM units [\texttt{LSTM}], a mixed Convolutional-Recurrent Neural Network [\texttt{CLSTM}] and Attention mechanism based Auto-encoder with Convolutional input layers [\texttt{ATT}].
\end{labeling}

All datasets, Python source code, details on the hyper-parameters sought for every model in the benchmark, sizes of Deep Learning models (number of trainable parameters), and simulation results are publicly available at \url{https://github.com/Eric-L-Manibardo/CaseStudy2020}.

\subsection{Results and statistical analysis}

The obtained simulation results are presented an analyzed hereby, emphasizing on the performance gaps between models and their statistical significance.

\begin{figure}[ht!]
    \centering
    \includegraphics[width=\columnwidth]{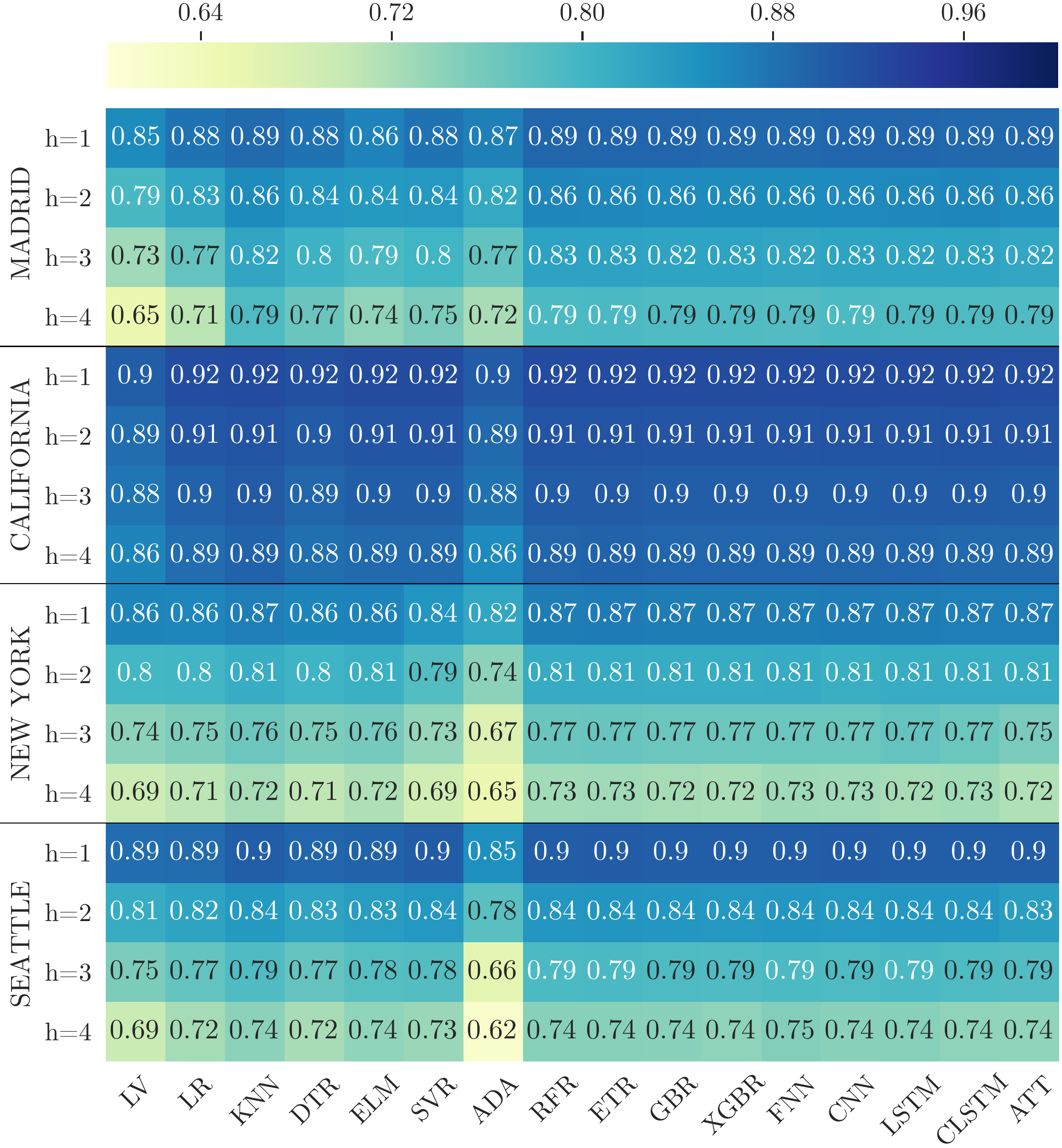}
    \caption{Heatmap showing the average $R^2$ test score obtained by each model and dataset. Values are computed as the mean value of the test scores obtained for the 10 locations selected for each data source and value of the forecasting horizon. Rows stand for data source and forecasting horizon, while columns correspond to the considered forecasting models.}
    \label{fig:heatmap}
\end{figure}

The discussion begins with Figure \ref{fig:heatmap}, which displays the overall performance, computed as the mean $R^2$ score averaged over the 10 datasets of each data source, for every learning method and analyzed forecasting horizon $h$. As expected, the performance of the models degrades consistently as the prediction horizon increases. Traffic data corresponding to the California data source are stable, which can be appreciated by a simple visual inspection of their profiles: a high $R^2$ score is obtained for this dataset even when predicting four steps ahead ($h=4$). As stated in Section \ref{subsec:review}, the PeMS data source is the most popular option for ITS studies, especially when novel forecasting methods are presented. In this study, we have collected only datasets from District 4 (the so-called \textit{Bay Area}), as data from other districts also provide stable traffic measurements, and District 4 is the most commonly selected sector among the revised literature. 

The nature of traffic measurements, jointly with the scope area of data sources, can suggest in advance how forecasting performance degrades when the prediction horizon $h$ is increased. Both in the city and in highways, drivers tend to maintain a nominal speed whenever possible, so time series drops suddenly. Thereby, only the last timestamps provide information on this phenomena \cite{manibardo2020new}. Results for New York and Seattle data sources corroborate this statement, where the performance degradation maintains a similarly decaying trend. In the case of flow data, traffic at urban roads can differ significantly depending on the selected location. Main roads maintain a nearly constant traffic flow as trucks, taxis, and other basic services vehicles occupy the roads at night and early morning hours. This is not the case of special districts like the surroundings of universities, shopping malls and recreational areas, which impact on the traffic flow trends according to the schedules of their activities. Traffic flow at highways does not face these issues, degrading the forecasting performance more smoothly when increasing the prediction horizon, as it can be observed in the California test results.

With the focus set on the results of each model for the same collection of datasets, some of them render similar scores. At a first glance, the five Deep Learning architectures under consideration perform similarly to ensemble methods (except \texttt{ADA}). Shallow learning methods obtained a slightly lower $R^2$ score. Nevertheless, if the payoff for a minor performance degradation is a faster training time and less computational resource requirements, shallow learning methods should be taken into consideration. \texttt{SVR} is an exception, which holds by far, the longest optimization time among the analyzed methods. As long as researchers do not set iteration limit when searching the hyper plane combination that best fits the data distribution, \texttt{SVR} can demand long hyper-parameter optimization periods \cite{smola2004tutorial}. To end with, the relatively good forecasting performance of the naive \texttt{LV} method for low values of the forecasting horizon $h$ imposes a narrow gap for improvement, as evinced by the negligible $R^2$ differences noted between models.

Given such small differences between the scores attained by the models, it is necessary to assess whether they are significant in the statistical sense. Traditionally standard null hypothesis testing has been adopted in this regard, including post-hoc tests and graphical representations (e.g. critical distance plots \cite{demvsar2006statistical}) to visually assess which counterparts in the benchmark are performing best with statistical significance. However, recently criticism has arisen around the use of these tests, due to their lack of interpretability and the sensitivity of their contributed statistical insights, and to the number of samples used for their computation. 

\begin{figure}[ht!]
    \centering
    \includegraphics[width=\columnwidth]{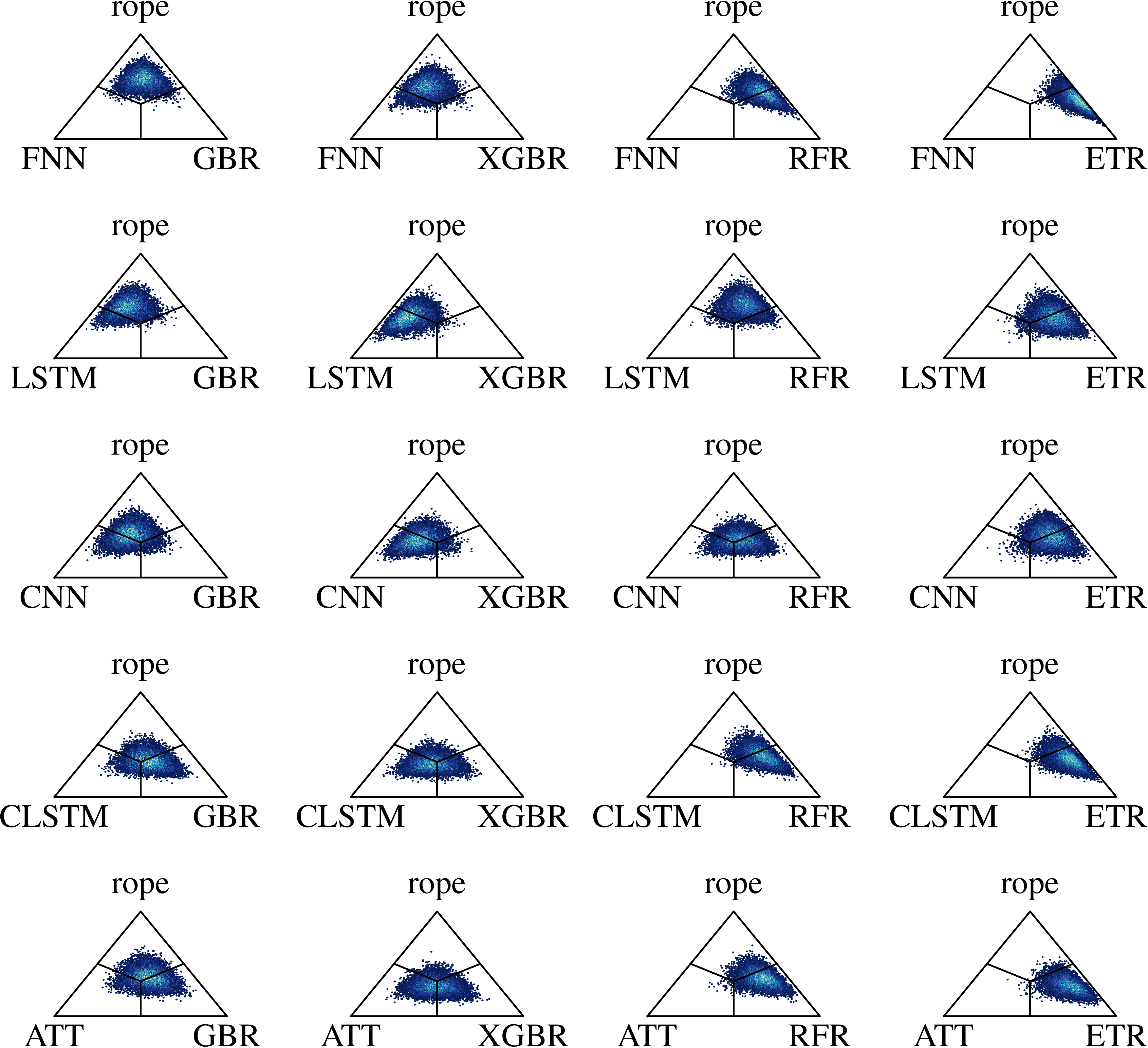}
    \caption{Bayesian probabilities sampled via Monte Carlo,  and $rope=0.001$ (i.e. absolute $R^2$ score differences under this value are considered to be equal). Rows correspond to Deep Learning models, whereas columns correspond to ensembles. Bright colors denote a higher probability of the fitted Gaussian distribution.}
    \label{fig:bayesian}
\end{figure}

In this context, the seminal work by Benavoli et al in \cite{benavoli2017time} exposed the drawbacks of standard hypothesis testing, and promoted the use of Bayesian analysis for multiple comparison analysis. We embrace this new methodological trend, and compute a Bayesian analysis between every (Deep Learning, ensemble) model pair, which output is shown in Figure \ref{fig:bayesian} (rows: Deep Learning models, columns: ensemble models). Bayesian analysis performed on every such pair allows computing the probability that one model outperforms another, based on the test results obtained by each of them over all locations, datasets and $h$ values. The obtained probability distribution can be sampled via Monte Carlo and displayed in barycentric coordinates, comprising two regions: one where the first model outperforms the second, and vice-versa. Additionally, a region of practical equivalence (where results can be considered to be statistically equivalent) can be set as per a parameter called $rope$. This parameter indicates the minimum difference between the scores of both methods for them to be considered significantly different to each other. The value of $rope$ depends on the task being solved. For example a forecasting error difference of one single car when predicting traffic flow at highways of 300 passing vehicles per analyzed interval can be ignored, as this margin does not affect a practical implementation of the predicting models.  

The results of the Bayesian analysis depicted in Figure \ref{fig:bayesian} reveals that \texttt{LSTM} and \texttt{CNN} have a slightly higher probability of providing better results than \texttt{GBR} and \texttt{XGBR} ensembles. However, the situation changes for \texttt{RFR} and \texttt{ETR}. The sampled probabilities of both ensembles when compared to Deep Learning variants are skewed towards the regions of practical equivalence (e.g. \texttt{RFR} versus \texttt{LSTM}) or towards the region where the ensemble performs better than the Deep Learning models (e.g. \texttt{ETR} versus \texttt{CLSTM}). On a concluding note, the statistical analysis concludes that from the statistical point of view, there is no clear winner in the benchmark, nor any empirically supported reason for using Deep Learning based traffic forecasting models detrimentally to shallow modeling alternatives. 

\section{Learned Lessons}\label{sec:learnedlessons}

It has been concluded from the experimental results that Deep Learning models do not provide consistently better results than shallow modeling approaches. Furthermore, whenever hyper-parameters are properly tuned beforehand, ensemble methods outperform Deep Learning models in some cases. This fact demonstrates that parameter tuning should be mandatory in prospective studies to avoid unfair comparisons. Unfortunately, hyper-parameter tuning stage is often neglected or mentioned very superficially, without the relevance it deserves. 

Besides, the training complexity of this kind of algorithms is widely overlooked. Our literature analysis unveils that short-term traffic forecasting  publications are leaning towards more complex models on the understanding that their increased modeling power can improve the state of the art, often by narrow performance margins. However, such slight performance gaps do not translate into practical advantages for real traffic scenarios \cite{bratsas2020comparison}. For a similar and sometimes even better result, classic Machine Learning techniques can perform as well as Deep Learning, but with less complexity and computational requirements.

It is also important to underscore the essential role of naive methods when establishing the minimum complexity of the designed task (Figure \ref{fig:heatmap}). These baseline models should take part in any traffic forecasting benchmark. The task to be solved in the case study (i.e. predicting traffic state at a single road) was chosen on purpose to show that for simple tasks, complex models do not significantly improve the performance of a naive model. The most meaningful information for the target to be predicted is made available at the input of every model (previous recent measurements collected at the target road). Consequently there are no complex relationships to be modeled, and ultimately, Deep Learning architectures can not provide better results than shallow learning methods. A lower performance bound can also be established by means of autoregressive models, but they are very sensitive to parameter configuration. By contrast, the lack of parameters of naive methods make them a better choice to ascertain the improvement margin that can be achieved by virtue of data-based models.

Another relevant aspect is how train and test data are arranged. A common practice observed in the literature is that test data are carefully chosen in order to obtain the desired performance for the presented traffic forecasting method. Test data are often selected from short temporal intervals, with almost identical characteristics than the training data. This methodology neglects some of the basic notions of Machine Learning: whenever possible, test data should be different (yet following the same distribution) than training data to check the generalization capabilities of the developed model. Some of the analyzed papers reserve only one month of traffic data for training, and one week for testing. As a result of this partitioning criterion, the results can be misleading as learned traffic behavior can be identical to that present in the test subset, thereby generalizing poorly when modeling traffic belonging to other periods along the year.

In this context, different train/test partitioning choices are enabled by the amount of available data. In the best of circumstances, the data source covers at least two complete years, so researchers can train the model over the data collected in the first year, and check its generalization capabilities by testing over the data of the second year. Throughout the year, the traffic profile can change in some points of a traffic network due to e.g. road adjustments, extreme meteorological events or sociopolitical decisions. These circumstances generate unusual traffic daily patterns that modify the data distribution, inducing an additional level of difficulty for the learning and adaptation capabilities of data-based models. In this context, it is remarkable the fact that PeMS, arguably the most commonly used data source as it provides several years of traffic measurements, is not commonly utilized over the entire time span covered by this dataset. 

The second option is to have only one complete year of traffic data. In this case, we suggest arranging the data as done in our case study: three weeks of every month as train data, and the remaining days of every month for testing. This configuration allows the model to learn from different traffic patterns, so that authors can check if the model generalizes properly to unseen data using the test holdout and considering, at least, all traffic behaviors that can occur during the year for the location at hand. 

The last case corresponds to a data source that does not cover an entire year. In this scenario, the generalization of the model's performance to the overall year cannot be fully guaranteed because, depending on the time range covered by the dataset, patterns learned by the model can only be used to produce forecasts for a short period of the year. Given the amount of traffic data available nowadays for experimentation, it should not be an issue for prospective works to find a public traffic data source that matches the desired characteristics for the study, and also provides at least a full year of measurements.

Finally, a good practice that unfortunately is not mostly adopted in traffic forecasting is to release the source code and data producing the results of the proposed model to the public domain. This practice would ease the revision process, ensure the reproducibility of the reported results, and foster future research efforts aimed at their improvement. Clearly, this practice is stringently subject to the confidentiality of the traffic data under consideration, but whenever possible, traffic datasets, source and results should be left in public repositories (e.g. GitHub, BitBucket and the like), so that new ideas and investigations do not depart from scratch, and advances over the field become more reliable, verifiable and expedited. 

\section{Challenges and Research Opportunities}\label{sec:challenges}

\begin{figure}[h!]
    \centering
    \includegraphics[width=\columnwidth]{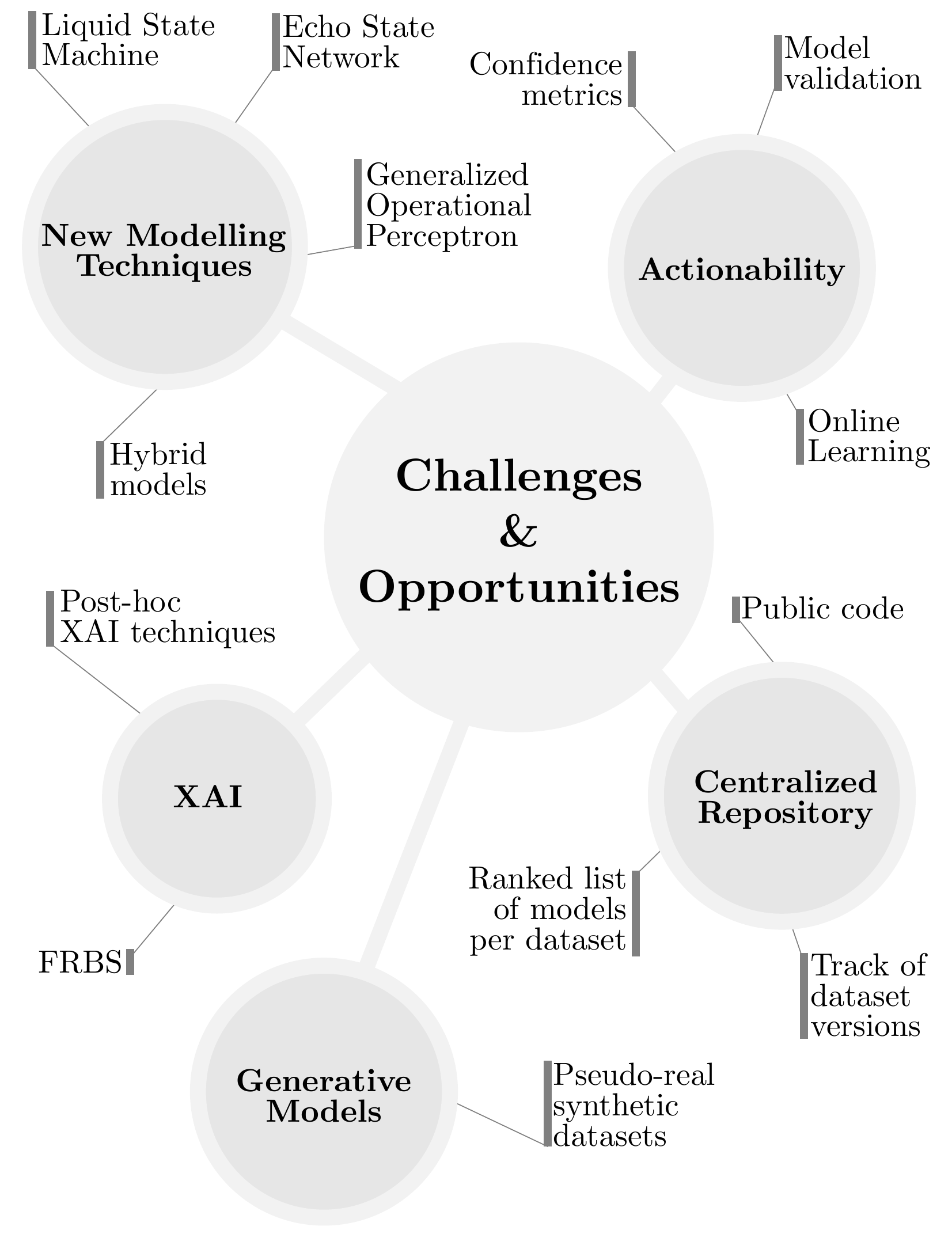}
    \caption{Schematic overview of identified challenges and suggested research opportunities. }
    \label{fig:challenges}
\end{figure}

As new data processing and modeling techniques flourish in the community, emerging research paths arise to yield more precise and wider covering traffic forecasting models. This section points out challenges that need to be faced, as well as research opportunities that should be explored by the community in years to come. Figure \ref{fig:challenges} summarizes graphically our vision on the future of this research area, which we next describe in detail.

\subsection{Actionability: adaptive models and prediction confidence}

The literature review has demonstrated that there is an increasing race towards finding the best performing traffic forecasting model. However, model \emph{actionability} should be the ultimate goal for works in the field, which has not exclusively to do with the precision of the forecasts \cite{lana2020data}. 

If we split data-driven modeling into sequential stages, a traffic forecasting scenario covers 1) data sensing; 2) data preprocessing, ending in built regression datasets; 3) a learning and validation phase, where a model is learned from such datasets; and 4) model testing, where the performance of the trained model is verified when predicting unseen traffic data. When one of these stages is granted too much relevance, important aspects in other phases of the data pipeline can be neglected. For instance, datasets are sometimes composed of handpicked locations of the traffic network (i.e. the data source), coincidentally those with more stable patterns that could lead to unrealistically good model performance levels. 

Additionally, traffic data might evolve over long time periods, which leads to the fifth and often overseen stage: model adaptation \cite{matias2015improving}. The idea of model adaptation is conceptually simple: traffic data is continuously fed to the model, which uses the new information to adapt to contextual changes affecting its learned knowledge \cite{buchanan2015traffic,pan2013crowd}. For this purpose, online learning techniques allow for the incremental update of the model when fed with new data, whereas concept drift handling approaches permit to adapt the behavior of the forecasting model to changing data distributions. Although the literature provides specific publications about this topic \cite{manibardo2020new, lana2019adaptive, mena2017improving, wu2012online,procopio2009learning}, it remains as a largely uncharted research area in traffic forecasting.

Lastly, for a model to become fully actionable, we firmly advocate for the addition of confidence metrics to predictions, so that traffic managers can trust and assess the uncertainty associated to the traffic forecasts, and thus make better informed decisions. From a strategic point of view, confidence estimation in travel demand prediction has a solid research background \cite{nicolaisen2014ex,parthasarathi2010post,yang2013sensitivity,rasouli2014using,welde2011planners}, which helps design and scale properly road infrastructure. Confidence for long-term congestion predictions have also relevant contributions \cite{lana2019question,matas2012traffic}. However, there are no remarkable contributions on this matter for short-term traffic forecasting. 

All in all, forecasting models are the bridge connecting raw data to reliable decisions for traffic management. This need for actionable decisions require far more insights that a single quantitative proof of the average precision achieved by forecasting models.

\subsection{Need for a centralized traffic data repository}

The review of selected works has uncovered an increasing number and diversity of traffic data sources in use during recent years. The issue arises precisely by the number of available options. Even for a specific data source, different datasets can be furnished depending on the location of measurement, time intervals or aggregation rate, among other choices. Researchers often apply different preprocessing techniques (usually designed and implemented ad-hoc for the study) to prepare the data for better modeling performance due to more representative examples. For this reason, the ITS community has so far generated multiple versions of many data sources, leading to incongruities in benchmarks comprising state of the art solutions. 

All these issues could be overcome if a single point of information was made available for the community: in short, a \emph{centralized traffic data repository}. This repository would store different versions of traffic datasets in an uniform format, according to the different preprocessing techniques applied to the original traffic data sources. The repository would also publish a ranked list of the best performing models for each dataset and forecasting task, for the sake of fair comparisons between novel models. Researchers could reference datasets from third-party research works, and compare their newly proposed technique to previous ones. Interfaces enabling the submission of new data-based pipelines, datasets and results would also be unleashed for extending the coverage of this repository, including the source code producing the results published in the corresponding publication.

Definitely, the availability of this centralized repository would accelerate the understanding of the current status of the field, favoring the development of new and more reliable model comparisons. We illustrate this idea by sharing the processed datasets employed during the case study of Section \ref{sec:casestudy} in a freely accessible GitHub repository. We firmly believe that the integration of our repository and others scattered over the literature into a single point of information will be a long awaited milestone for the community working in traffic forecasting.

\subsection{Generative models for pseudo-real synthetic datasets }

The vast majority of learning methods selected by the ITS community attempt to model the conditional probability $P(y \vert \emph{\textbf{x}})$, where the desired output value $y$ (e.g. the traffic forecast) is conditioned by the input \emph{\textbf{x}} (the predictor variables at the input of the forecasting model). On the other hand, \emph{generative models} estimate $P(\emph{\textbf{x}} \vert y)$, as they try to learn the conditional distribution of data \cite{xue2008comment}. As their name suggests, these models can generate new synthetic data instances, opening an interesting research path towards augmenting the amount of traffic data with which models are trained.

Although researchers have access to traffic simulators like CORSIM \cite{halati1997corsim}, VISSIM \cite{fellendorf2010microscopic}, or SUMO  \cite{behrisch2011sumo}, these tools serve a specific purpose: to provide simulated traffic environments with a concrete collection of features. Here, the fictional traffic network is designed and shaped by selecting parameters such as the number of vehicles, speed, road design, etc. Due to this tuning, the environment is conditioned by the investigation requirements and lose its realistic nature. On this line, \emph{generative models} could provide synthetic data, that resemble real traffic networks. With this, scarce data sources from key locations could be extended, for scenarios where test holdout does not cover all possible traffic states.

In particular, Generative Adversarial Networks (GANs) \cite{goodfellow2016nips} have demonstrated notable results at learning to synthesize new data instances that highly resemble real data. There are hundreds of publications reported in recent times using GANs for spatio-temporal data \cite{gao2020generative}. We foresee that these generative models will acquire a capital importance in traffic forecasting, especially in traffic forecasting scenarios with scarce data. Some recent achievements have already showcased the potential of GANs for this purpose \cite{saxena2019d,luo2018multivariate}, paving the way towards massively incorporating these models for traffic forecasting under data availability constraints.

\subsection{New modeling techniques for traffic forecasting}

Another research path garnering a significant interest in recent times aims at the application of alternative data-based modeling approaches to traffic forecasting, mainly towards advancing over the state of the art in terms of design factors beyond the precision of their produced predictions (e.g. computational complexity of the underlying training process). This is the case of recent attempts at incorporating elements from Reservoir Computing \cite{lukovsevivcius2009reservoir} and randomization-based Machine Learning to the traffic prediction realm, including echo state networks \cite{yang2012short}, extreme learning machines \cite{lou2020probabilistic}, or more elaborated variants of these modeling alternatives \cite{del2019road,del2020deep}. The extremely efficient learning procedure of these models makes them particularly appropriate for traffic forecasting over large datasets. On the other hand, the high parametric sensitivity of models currently utilized for traffic forecasting has also motivated the renaissance of bagging and boosting tree ensembles for the purpose, which are known to be more robust against the variability of their hyper-parameters and less prone to overfitting \cite{yang2017ensemble,lu2020short,li2020short}. Finally, initial evidences of the applicability of automated machine learning tools for efficiently finding precise traffic forecasting models have been recently reported in \cite{angarita2020evaluating}.

All in all, there is little doubt that most discoveries and innovations in data-based modeling are nowadays related to Deep Learning. However, beyond the lessons and good practices exposed previously for embracing their use, we advocate for a closer look taken at other modern modeling choices, such as the Generalized Operational Perceptron \cite{tran2019heterogeneous}, Liquid State Machines \cite{maass2011liquid}, or models encompassing an hybridization of traffic flow models and machine learning techniques \cite{zhang2020hybrid}. Likewise, other design objectives that do not relate strictly to the accuracy of issued predictions should be increasingly set under target, mostly considering the huge scales attained today by traffic data. A major shift towards efficiency is needed for data-based traffic forecasting models, making use of new evaluation metrics that take into account the amount of data and/or number of operations required for model training.

\subsection{Understanding and explaining Deep Learning models}

When trained, Deep Learning models are black-boxes that do not grant any chance for the general user to understand how their predictions are made \cite{arrieta2020explainable,adadi2018peeking}. In the case of traffic operators, the reasons why a neural network produces a particular prediction are of utmost necessity for making informed decisions. In a situation of disagreement, in which the operator of the traffic network does not trust the model prediction, Deep Learning does not offer any means to explain the captured knowledge that led to its forecasts. Similarly to other fields of knowledge (e.g. medical diagnosis), this lack of transparency of Deep Learning models makes it hard for humans to accept their predictions, who often opt for worse performing yet transparent alternatives (e.g. regression trees). 

To the best of our knowledge, very few publications have tackled traffic forecasting from a eXplainable Artificial Intelligence (\emph{XAI}) perspective. One example is \cite{sun2006bayesian}, which studies the cause-effect relationship between nodes of a traffic network, attempting at learning how upstream and downstream traffic influence the traffic prediction at the target road. A model based on a stacked auto-encoder for missing and corrupt data imputation is presented in \cite{duan2016efficient}, where the features extracted by the first hidden layer are analyzed towards improving the interpretability of model decisions. In \cite{wu2018hybrid}, authors develop an attention-based traffic forecasting model. Then, for a better understanding of the propagation mechanism learned by the model, they examine the evolution of these attention scores with respect to spatial and temporal input data. The last example is \cite{barredo2019lies}, where knowledge from two surrounding roads is studied by analyzing the importance of the traffic features (i.e. flow values from different time steps of the time series from these roads) by using a post-hoc XAI technique.

Most cause-effect relationships in traffic data are studied theoretically \cite{kerner1999congested,treiber1999explanation}, without considering the complexity that comes from the use of Deep Learning techniques. Even with correct predictions, a model that is not understandable can be of no practical value for traffic managers willing to obtain insights beyond its predicted output. In recent years, the family of Fuzzy Rule Based Systems (FRBS) model has experienced a renaissance thanks to their envisaged relevance within the XAI paradigm \cite{fernandez2019evolutionary}. FRBS learn a set of human-readable \emph{if then} rules defined on a fuzzy domain that best correlate the predictors and the target variable. We envision that these models, along with post-hoc XAI techniques specific to Deep Learning models, will be central for the acceptance of shallow and Deep Learning models in traffic management processes. Specifically, fuzzy rules built for explaining the knowledge captured by  black-boxes, and other forms for visualizing local explanations of the produced forecasts will surely contribute to their use in practical deployments, further contributing to the actionability of their issued predictions.

\section{Conclusions}\label{sec:conclusions}

This critical survey has departed from the abundance of contributions dealing with Deep Learning techniques for road traffic forecasting. In the mid 80's, the community began to model traffic distributions using data-driven methods, replacing statistical approaches prevailing at the time. Years thereafter, Deep Learning based models has taken the lead in the field, spurred by the unprecedented performance increases observed in other application domains. Their renowned superior modeling capability made the community steer towards Deep Learning based traffic forecasting models, yet without pausing and profoundly reflecting on their benefits and downsides. 
Our literature review, which comprises more than 150 works at the crossroads between Deep Learning and short-term traffic forecasting, has revealed the lights and shadows of the current state of this research area. As a result, we have identified a number of questionable methodological practices and points of improvement, prescribing a set of recommendations of future studies:
\begin{itemize}[leftmargin=*]
\item An adequate selection of traffic datasets, complemented by a preprocessing stage that yields properly partitioned train and test subsets.

\item An appropriate reasoning of the choice of Deep Learning problems, supported by the need for fusing heterogeneous contextual and/or spatio-temporal data.

\item A  principled comparison study, encompassing baseline models, different metrics beyond precision (e.g. computational efficiency) and a statistical study aimed at concluding whether the metric gaps are statistically significant.
\end{itemize}

To further clarify whether Deep Learning makes a difference in traffic forecasting, we have designed a case study intended to serve as an argument for our claims. The obtained results render empirical evidence about two main facts: 1) the nature and scope of the selected traffic dataset establishes the complexity of the forecasting task, so challenging traffic datasets are recommended for model comparison purposes; and 2) when choosing a time series regression model for traffic forecasting, Deep Learning provides similar performance levels than shallow learning models, or at least no statistically better whatsoever. We have summarized our conclusions as a set of learned lessons, which sets forth good practices for future short-term traffic forecasting studies. 

Our overall analysis ends up with an outlook on the challenges that persist unaddressed in the ITS field in what refers to traffic forecasting. Research opportunities are also given for approaching such challenges, partly inspired by recent achievements in data-based modelling. Among them, we have highlighted the need for taking a step further beyond accuracy, to account for other aspects that favor the actionability of traffic forecasts (e.g. confidence estimation and model explainability). Besides, we envision that a centralized traffic data repository would allow researchers to use the same traffic datasets and to reproduce results reported in the literature. Finally, the use of generative models for creating realistic traffic data will span further opportunities for data augmentation.

Despite our constructive criticism exposed throughout the paper, we agree on the flexibility that makes Deep Learning excel at modeling diverse phenomena and outperforming other data-driven models. However, it is our belief that as in other disciplines, the adoption of Deep Learning for traffic forecasting should be grounded on a fair assessment of the benefits and drawbacks it may yield. Our experiments have proven that shallow learning methods provide similar results when compared to Deep Learning architectures at a lower computational complexity, whenever comparisons are done in a principled manner. Nevertheless, far from proposing to leave it aside, we firmly defend that Deep Learning should be embraced only when its singular capabilities provide performance gains worth the extra computational cost.

\section*{Acknowledgments}

The authors would like to thank the Basque Government for its funding support through the EMAITEK and ELKARTEK programs (3KIA project, KK-2020/00049). Eric L. Manibardo receives funding support from the Basque Government through its BIKAINTEK PhD support program (grant no. 48AFW22019-00002). Javier Del Ser also thanks the same institution for the funding support received through the consolidated research group MATHMODE (ref. T1294-19).

\bibliographystyle{IEEEtran}
\bibliography{DDD}

\begin{IEEEbiography}[{\includegraphics[width=1in,clip,keepaspectratio]{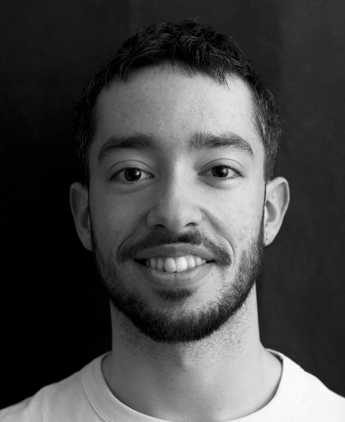}}]{Eric L. Manibardo} received his B.Sc. degree in Telecommunication Engineering in 2017, and M.Sc. degree also in Telecommunications Engineering in 2019 from the University of the Basque Country, Spain. He is currently a junior researcher at TECNALIA (Spain), pursuing his PhD in Artificial Intelligence. His research interest combine machine learning and signal processing within the context of Intelligent Transportation Systems (ITS), with an emphasis on traffic forecasting. 
\end{IEEEbiography}

\begin{IEEEbiography}[{\includegraphics[width=1in,clip,keepaspectratio]{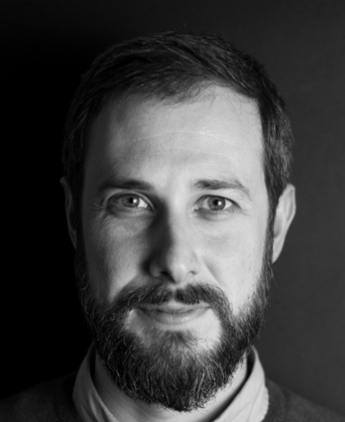}}]{Ibai La\~na} received his B.Sc. degree in Computer Engineering from Deusto University, Spain, in 2006, the M.Sc. degree in Advanced Artificial Intelligence from UNED, Spain, in 2014, and the PhD in Artificial Intelligence from the University of the Basque Country in 2018. He is currently a senior researcher at TECNALIA (Spain).  His research interests fall within the intersection of Intelligent Transportation Systems (ITS), machine learning, traffic data analysis and data science. He has dealt with urban traffic forecasting problems, where he has applied machine learning models and evolutionary algorithms to obtain longer term and more accurate predictions. He is currently researching methods to measure the confidence of traffic and other time series data. He also has interest in other traffic related challenges, such as origin-destination matrix estimation or point of interest and trajectory detection.
\end{IEEEbiography}

\begin{IEEEbiography}[{\includegraphics[width=1in,height=1.25in,clip,keepaspectratio]{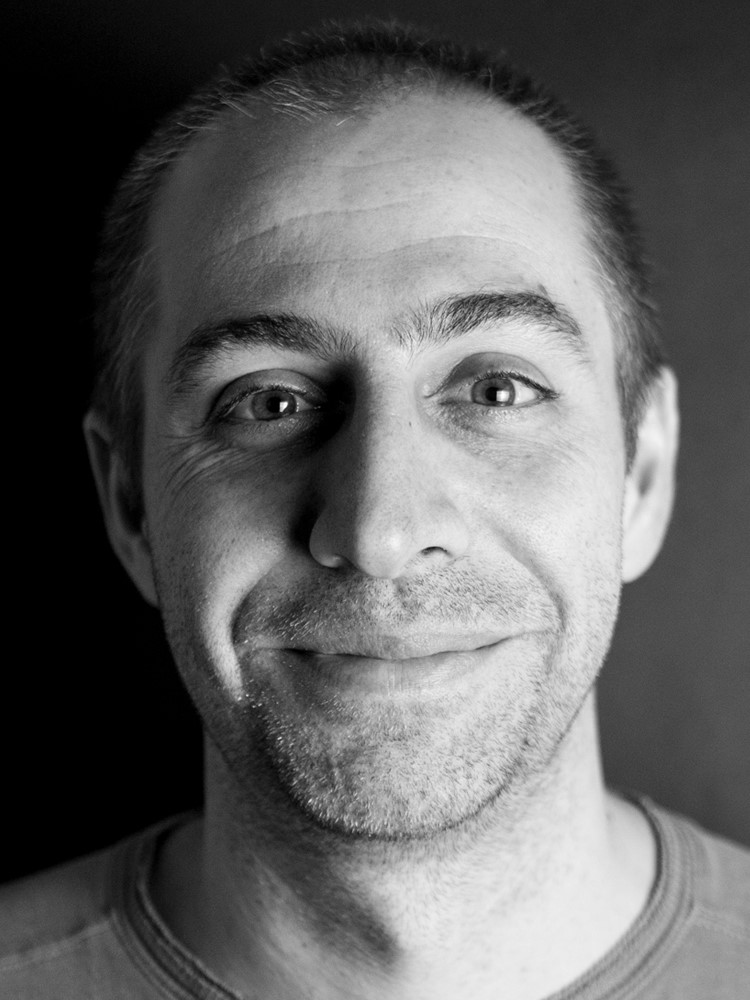}}]{Javier Del Ser} (SM'12) received his first PhD degree (cum laude) in Electrical Engineering from the University of Navarra (Spain) in 2006, and a second PhD degree (cum laude, extraordinary PhD prize) in Computational Intelligence from the University of Alcala (Spain) in 2013. He is currently a Research Professor in Artificial Intelligence and leading scientist of the OPTIMA (Optimization, Modeling and Analytics) research area at TECNALIA, Spain. He is also an adjunct professor at the University of the Basque Country (UPV/EHU), and an invited research fellow at the Basque Center for Applied Mathematics (BCAM). His research interests are in the design of Artificial Intelligence methods for data mining and optimization applied to problems emerging from Intelligent Transportation Systems, Smart Mobility, Logistics and Autonomous Driving, among specific interests in other domains. He has published more than 380 scientific articles, co-supervised 10 Ph.D. theses, edited 7 books, co-authored 9 patents and participated/led more than 40 research projects. He is an Associate Editor of tier-one journals from areas related to Artificial Intelligence, such as Information Fusion, Swarm and Evolutionary Computation and Cognitive Computation, as well as an Associate Editor of IEEE Transactions on Intelligent Transportation Systems.
\end{IEEEbiography}
\vfill
\end{document}